\documentclass[nojss]{jss}\usepackage[]{graphicx}\usepackage[]{color}
\makeatletter
\def\maxwidth{ %
  \ifdim\Gin@nat@width>\linewidth
    \linewidth
  \else
    \Gin@nat@width
  \fi
}
\makeatother

\definecolor{fgcolor}{rgb}{0.345, 0.345, 0.345}
\newcommand{\hlnum}[1]{\textcolor[rgb]{0.686,0.059,0.569}{#1}}%
\newcommand{\hlstr}[1]{\textcolor[rgb]{0.192,0.494,0.8}{#1}}%
\newcommand{\hlopt}[1]{\textcolor[rgb]{0,0,0}{#1}}%
\newcommand{\hlstd}[1]{\textcolor[rgb]{0.345,0.345,0.345}{#1}}%
\newcommand{\hlkwb}[1]{\textcolor[rgb]{0.69,0.353,0.396}{#1}}%
\newcommand{\hlkwc}[1]{\textcolor[rgb]{0.333,0.667,0.333}{#1}}%
\newcommand{\hlkwd}[1]{\textcolor[rgb]{0.737,0.353,0.396}{\textbf{#1}}}%

\usepackage{framed}
\makeatletter
\newenvironment{kframe}{%
 \def\at@end@of@kframe{}%
 \ifinner\ifhmode%
  \def\at@end@of@kframe{\end{minipage}}%
  \begin{minipage}{\columnwidth}%
 \fi\fi%
 \def\FrameCommand##1{\hskip\@totalleftmargin \hskip-\fboxsep
 \colorbox{shadecolor}{##1}\hskip-\fboxsep
     \hskip-\linewidth \hskip-\@totalleftmargin \hskip\columnwidth}%
 \MakeFramed {\advance\hsize-\width
   \@totalleftmargin\z@ \linewidth\hsize
   \@setminipage}}%
 {\par\unskip\endMakeFramed%
 \at@end@of@kframe}
\makeatother

\definecolor{shadecolor}{rgb}{.97, .97, .97}
\definecolor{messagecolor}{rgb}{0, 0, 0}
\definecolor{warningcolor}{rgb}{1, 0, 1}
\definecolor{errorcolor}{rgb}{1, 0, 0}
\newenvironment{knitrout}{}{} 

\usepackage{alltt}

\usepackage{stmaryrd}
\usepackage[usenames,dvipsnames,svgnames,table]{xcolor}
\usepackage{amsmath}

\definecolor{mygrey}{RGB}{150, 150, 150}

\author{Pascal Kerschke\\Information Systems and Statistics\\University of M{\"u}nster}
\title{Comprehensive Feature-Based Landscape Analysis of Continuous and Constrained Optimization Problems Using the \proglang{R}-Package \pkg{flacco}}

\Plainauthor{Pascal Kerschke} 
\Plaintitle{Comprehensive Feature-Based Landscape Analysis of Continuous and Constrained Optimization Problems Using the R-Package flacco} 
\Shorttitle{Feature-Based Landscape Analysis of Continuous Optimization Problems Using \pkg{flacco}} 

\Abstract{
This paper provides an introduction into the \proglang{R}-package \pkg{flacco}. A slightly modified version of this document has recently been submitted to the \href{https://www.jstatsoft.org/index}{\textit{Journal of Statistical Software}} and is currently under review.

Choosing the best-performing optimizer(s) out of a portfolio of optimization algorithms is usually a difficult and complex task. It gets even worse, if the underlying functions are unknown, i.e.,~so-called \textit{Black-Box problems}, and function evaluations are considered to be expensive. In the case of continuous single-objective optimization problems, \textit{Exploratory Landscape Analysis (ELA)} -- a sophisticated and effective approach for characterizing the landscapes of such problems by means of numerical values before actually performing the optimization task itself -- is advantageous. Unfortunately, until now it has been quite complicated to compute multiple ELA features simultaneously, as the corresponding code has been -- if at all -- spread across multiple platforms or at least across several packages within these platforms.

This article presents a broad summary of existing ELA approaches and  introduces \pkg{flacco}, an \proglang{R}-package for \textbf{f}eature-based \textbf{l}and\-scape \textbf{a}nalysis of \textbf{c}ontinuous and \textbf{c}onstrained \textbf{o}ptimization problems. Although its functions neither solve the optimization problem itself nor the related \textit{Algorithm Selection Problem (ASP)}, it offers easy access to an essential ingredient of the ASP by providing a wide collection of ELA features on a single platform -- even within a single package. In addition, \pkg{flacco} provides multiple visualization techniques, which enhance the understanding of some of these numerical features, and thereby make certain landscape properties more comprehensible. On top of that, we will introduce the package's build-in, as well as web-hosted and hence platform-independent, graphical user interface (GUI), which facilitates the usage of the package -- especially for people who are not familiar with \proglang{R} -- making it a very convenient toolbox when working towards algorithm selection of continuous single-objective optimization problems.

}
\Keywords{Exploratory Landscape Analysis, \proglang{R}-Package, Graphical User Interface, Single-Objective Optimization, Continuous Optimization, Algorithm Selection, Black-Box Optimization}
\Plainkeywords{Exploratory Landscape Analysis, R-Package, Graphical User Interface, Single-Objective Optimization, Continuous Optimization, Algorithm Selection, Black-Box Optimization} 

\Address{
  Pascal Kerschke\\
  Information Systems and Statistics\\
  University of M{\"u}nster\\
  Leonardo-Campus 3\\
  48149 M{\"u}nster, Germany\\
  E-mail: \email{kerschke@uni-muenster.de}\\
  URL: \url{http://erc.is/p/kerschke}
}

\IfFileExists{upquote.sty}{\usepackage{upquote}}{}
\begin{document}
\shortcites{Mersmann2011, Mersmann2013, Kerschke2014, Mlr2016, Plotly2016}


\newpage
\section{Introduction}\label{sec:intro}
Already a long time ago, \cite{Rice1976} introduced the \textit{Algorithm Selection Problem (ASP)}, which aims at finding the best algorithm $A$ out of a set of algorithms $\mathcal{A}$ for a specific instance\footnote{In this context, an \textit{instance} is the equivalent to an optimization problem, i.e.,~it maps the elements of the decision space $\mathcal{X}$ to the objective space $\mathcal{Y}$.} $I \in \mathcal{I}$. Thus, one can say that the algorithm selection model $m$ tries to find the best-performing algorithm $A$ for a given set of problem instances $\mathcal{I}$.

\begin{figure*}[b]
  \centering
  \includegraphics[width = \textwidth]{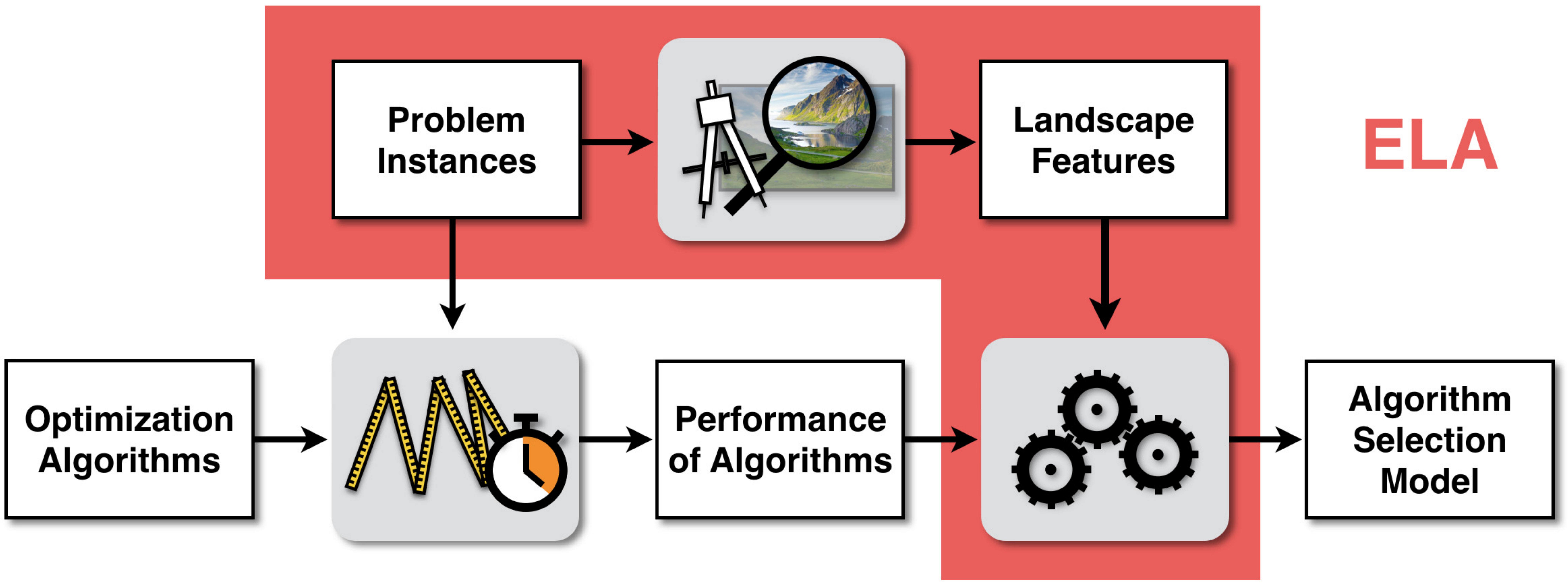}
  \caption{Scheme for using ELA in the context of an Algorithm Selection Problem.}
  \label{fig:ela_asp}
\end{figure*}

\begin{figure*}[!t]
  \centering
  \includegraphics[width = \textwidth, trim = 0mm 2mm 0mm 2mm, clip]{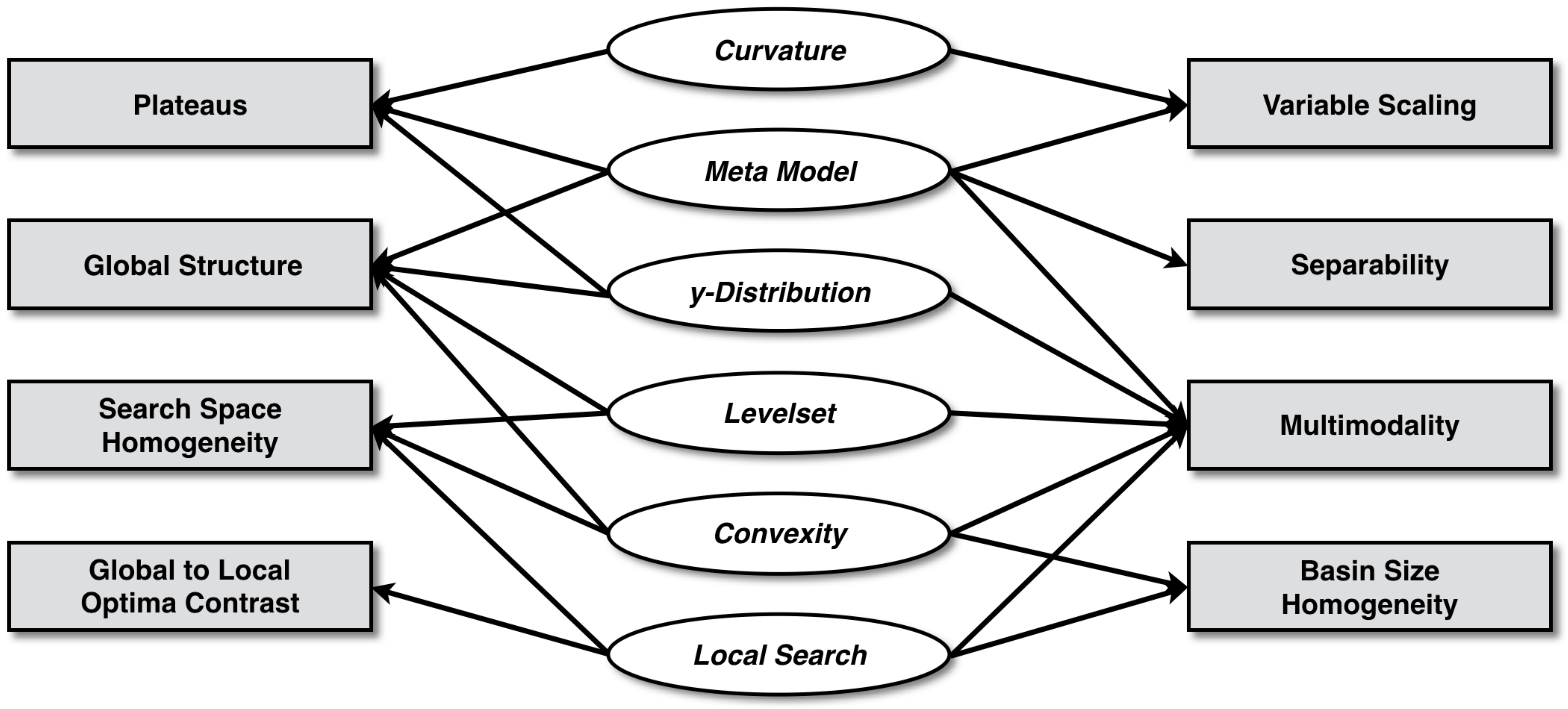}
  \caption{Relationship between expert-based high-level properties (grey rectangles in the outer columns) and automatically computable low-level properties (white ellipses in the middle).}
  \label{figure:elaprop}
\end{figure*}

As the performance of an optimization algorithm strongly depends on the structure of the underlying instance, it is crucial to have some a priori knowledge of the problem's landscape in order to pick a well-performing optimizer. One promising approach for such heads-up information are landscape features, which characterize certain landscape properties by means of numerical values.
Figure~\ref{fig:ela_asp} schematically shows how \textit{Exploratory Landscape Analysis (ELA)} can be integrated into the model-finding process of an ASP. In the context of continuous single-objective optimization, relevant problem characteristics cover information such as the degree of \textit{Multimodality}, its \textit{Separability} or its \textit{Variable Scaling}~\citep{Mersmann2011}. However, the majority of these so-called ``high-level properties'' have a few drawbacks: they (1) are categorical, (2) require expert knowledge (i.e.,~a decision maker), (3) might miss relevant information, and (4) require knowledge of the entire (often unknown) problem. Therefore, \cite{Bischl2012} introduced so-called ``low-level properties'', which characterize a problem instance by means of -- not necessarily intuitively understandable -- numerical values, based on a sample of observations from the underlying problem. Figure~\ref{figure:elaprop} shows which of their automatically computable low-level properties (shown as white ellipses) could potentially describe the expert-designed high-level properties (grey boxes).

In recent years, researchers all over the world have developed further (low-level) landscape features and implemented them in many different programming languages -- mainly in \proglang{MATLAB}~\citep{Matlab2013}, \proglang{Python}~\citep{Python2015} or \proglang{R}~\citep{R2017}. Most of these features are very cheap to compute as they only require an initial design, i.e.,~a small amount of (often randomly chosen) exemplary observations, which are used as representatives of the original problem instance.

For instance, \cite{Jones1995fitness} characterized problem landscapes using the \textit{correlation} between the distances from a set of points to the global optimum and their respective fitness values. Later, \cite{Lunacek2006} introduced \textit{dispersion} features, which compare pairwise distances of all points in the initial design with the pairwise distances of the best points in the initial design (according to the corresponding objective). \cite{Malan2009} designed features that quantify the \textit{ruggedness} of a continuous landscape and \cite{Mueller2011} characterized landscapes by performing \textit{fitness-distance analyses}. In other works, features were constructed using the problem definition, hill climbing characteristics and a set of random points \citep{Abell2013}, or based on a tour across the problem's landscape, using the change in the objective values of neighboring points to measure the landscape's \textit{information content} \citep{Munoz2012, Munoz2015}.

The fact that feature-based landscape analysis also exists in other domains -- e.g.,~in discrete optimization~\citep{Daolio2016, Jones1995, Ochoa2014} including its subdomain, the \textit{Traveling Salesperson Problem}~\citep{Mersmann2013, Hutter2014, Pihera2014} -- also lead to attempts to discretize the continuous problems and afterwards use a so-called \textit{cell mapping} approach~\citep{Kerschke2014}, or compute \textit{barrier trees} on the discretized landscapes~\citep{Flamm2002}.

In recent years, \cite{Morgan2015} analyzed optimization problems by means of \textit{length scale} features, \cite{Kerschke2015,Kerschke2016} distinguished between so-called ``funnel'' and ``non-funnel'' structures with the help of \textit{nearest better clustering} features, \cite{Malan2015} used \textit{violation landscapes} to characterize constrained continuous optimization problems, and \cite{Shirakawa2016} introduced the \textit{bag of local landscape features}.

Looking at the list from above, one can see that there exists a plethora of approaches for characterizing the fitness landscapes of continuous optimization problems, which makes it difficult to keep track of all the new approaches. For this purpose, we would like to refer the interested reader to the survey paper by \cite{Munoz2015_AS}, which provides a nice introduction and overview on various methods and challenges of this whole research field, i.e.,~the combination of feature-based landscape analysis and algorithm selection for continuous black-box optimization problems.

Given the studies from above, each of them provided new feature sets, describing certain aspects or properties of a problem. However, due to the fact that one would have to run the experiments across several platforms, usually only a few feature sets were combined -- if at all -- within a single study. As a consequence, the full potential of feature-based landscape analysis could not be exploited. This issue has now been addressed with the implementation of the \proglang{R}-package \pkg{flacco}~\citep{Flacco2017}, which is introduced within this paper.

While the general idea of \pkg{flacco}~\citep{Kerschke2016_CEC}, as well as its associated graphical user interface~\citep{Hanster2017} were already published individually in earlier works, this paper combines them for the first time and -- more importantly -- clearly enhances them by providing a detailed description of the implemented landscape features.

The following section provides a very detailed overview of the integrated landscape features and the additional functionalities of \pkg{flacco}. Section~\ref{sec:examples} demonstrates the usage of the \proglang{R}-package based on a well-known Black-Box optimization problem and Section~\ref{sec:visual} introduces the built-in visualization techniques. In Section~\ref{sec:gui}, the functionalities and usage of the package's graphical user interface is presented and Section~\ref{sec:summary} concludes this work.


\section{Integrated ELA features}\label{sec:ela}

The here presented \proglang{R}-package provides a unified interface to a collection of the majority of feature sets from above within a single package\footnote{The author also intends to extend \pkg{flacco} by the feature sets, which have not yet been integrated in it.} and consequently simplifies their accessibility. In its default settings, \pkg{flacco} computes more than 300 different numerical landscape features, distributed across 17 so-called \emph{feature sets} (further details on these sets are given on page~\pageref{item:featsets}). Additionally, the total number of performed function evaluations, as well as the required runtime for the computation of a feature set, are automatically tracked per feature set.

Within \pkg{flacco}, basically all feature computations and visualizations are based on a so-called \textit{feature object}, which stores all the relevant information of the problem instance. It contains the initial design, i.e.,~a data frame of all the (exemplary) observations from the decision space along with their corresponding objective values and -- if provided -- the number of blocks per input dimension (e.g.,~required for the \textit{cell mapping} approach,~\cite{Kerschke2014}), as well as the exact (mathematical) function definition, which is needed for those feature sets, which perform additional function evaluations, e.g.,~the \textit{local search} features (which were already mentioned in Figure~\ref{figure:elaprop}).

\begin{figure*}[!t]
  \centering
  \includegraphics[width = 0.9\textwidth]{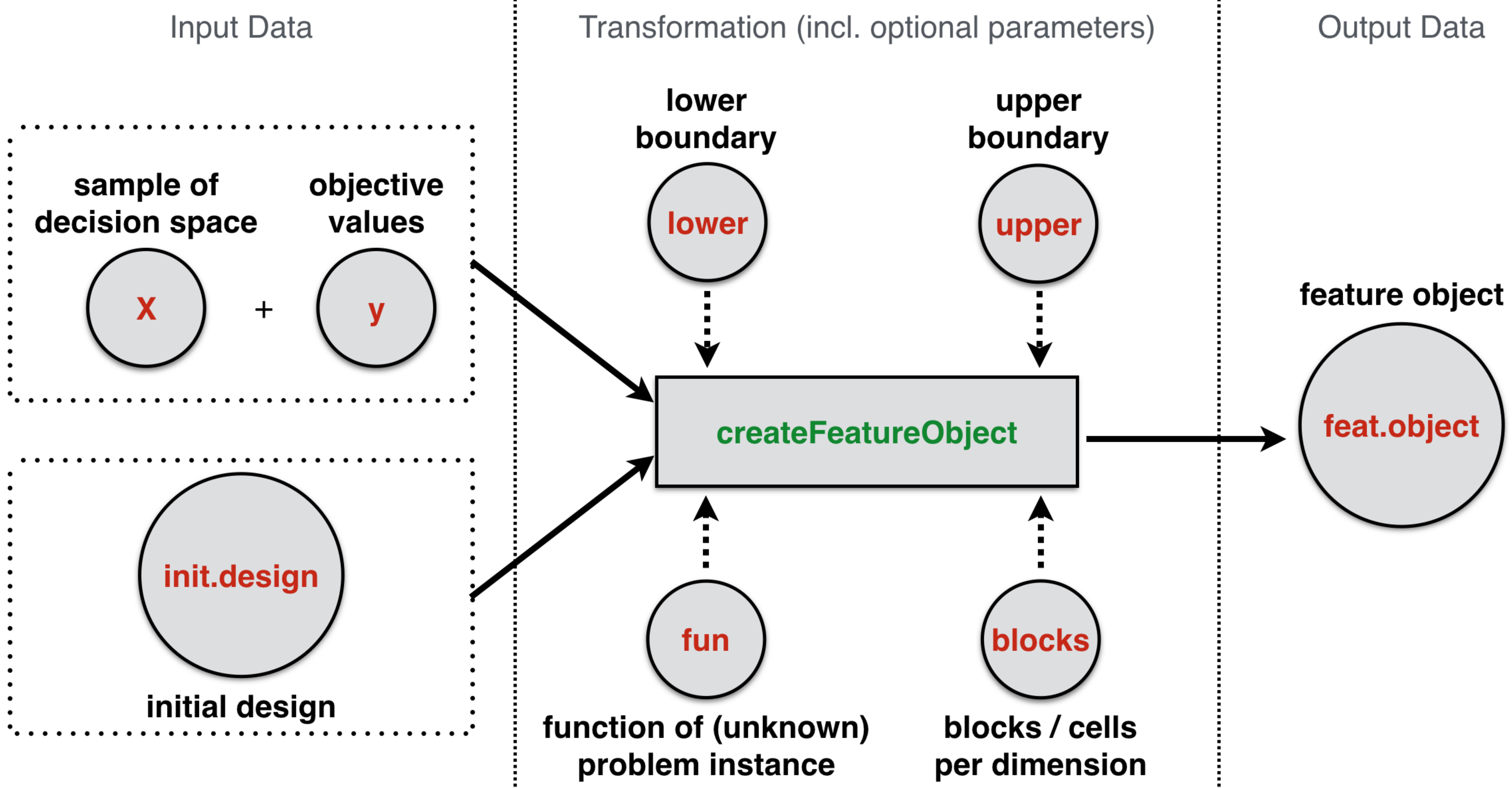}
  \caption{Scheme for creating a feature object: (1) provide the input data, (2) call \texttt{createFeatureObject} and pass additional arguments if applicable, (3) receive the feature object, i.e.,~a specific R-object, which is the basis for all further computations and visualizations.}
  \label{figure:featobject}
\end{figure*}

Such a feature object can be created in different ways as shown in Figure~\ref{figure:featobject}. First, one has to provide some input data by either passing (a) a matrix or data frame of the (sampled) observations from the decision space (denoted as \texttt{X}), as well as a vector with the corresponding objective values (\texttt{y}), or (b) a data frame of the initial design (\texttt{init.design}), which basically combines the two ingredients from (a), i.e.,~the decision space samples and the corresponding objective values, within a single data set. Then, the function \texttt{createFeatureObject} takes the provided input data and transforms it into a specific \proglang{R} object -- the feature object. During its creation, this object can (optionally) be enhanced / specified by further details such as the exact function of the problem (\texttt{fun}), the number of blocks per dimension (\texttt{blocks}), or the upper (\texttt{upper}) and lower (\texttt{lower}) bounds of the problem's decision space. If the latter are not provided, the initial design's minimum and maximum (per input variable) will be used as boundaries.

Furthermore, the package also provides the function \texttt{createInitialSample}, which allows to create a random sample of observations (denoted as \texttt{X} within Figure~\ref{figure:featobject}) within the box-constraints $[0, 1]^d$ (with $d$ being the number of input dimensions). This sample-generator also allows to configure the boundaries or use an \textit{improved latin hypercube sample}~\citep{Beachkofski2002, McKay2000} instead of a sample that is based on a random uniform distribution.

As one of the main purposes of \textit{Exploratory Landscape Analysis} is the description of landscapes when given a very restricted budget (of function evaluations), it is highly recommended to keep the sample sizes small. Within recent work, \cite{Kerschke2016} have shown that initial designs consisting of $50 \cdot d$ observations -- i.e.,~sampled points in the decision space -- can be completely sufficient for describing certain high-level properties by means of numerical features. Also, one should note that the majority of research within the field of continuous single-objective optimization deals with at most 40-dimensional problems. Thus, unless one possesses a high budget of function evaluations, as well as the computational resources, it is not useful to compute features based on initial designs with more than $10\,000$ observations or more than 40 features. And even in case one has those resources, one should consider whether it is necessary to perform feature-based algorithm selection; depending on the problems, algorithms and resources it might as well be possible to simply run all optimization algorithms on all problem instances.

Given the aforementioned feature object, one has provided the basis for calculating specific feature sets (using the function \texttt{calculateFeatureSet}), computing all available feature sets at once (\texttt{calculateFeatures}) or visualizing certain characteristics (described in more detail in Section~\ref{sec:visual}).

In order to get an overview of all currently implemented feature sets, one can call the function \texttt{listAvailableFeatureSets}. This function, as well as \texttt{calculateFeatures}, i.e.,~the function for calculating all feature sets at once, allows to turn off specific feature sets, such as those, which require additional function evaluations or follow the \textit{cell mapping} approach. Turning off the \textit{cell mapping} features could for instance be useful in case of higher dimensional problems due to the \textit{Curse of Dimensionality} \citep{Friedman1997}\footnote{In case of a 10-dimensional problem in which each input variable is discretized by three blocks, one already needs $3^{10} = 59\,049$ observations to have one observation per cell -- on average.}. 

In general, the current version of \pkg{flacco} consists of the feature sets that are described below. Note that the final two {``}features{''} per feature set always provide its costs, i.e., the additionally performed function evaluations and running time (in seconds) for calculating that specific set.

\paragraph{Classical ELA features (83 features across 6 feature sets):}\label{item:featsets}
The first six groups of features are the ``original'' ELA features as defined by \cite{Mersmann2011}.

Six features measure the \textit{convexity} by taking (multiple times) two points of the initial design and comparing the convex combination of their objective values with the objective value of their convex combinations. Obviously, each convex combination requires an additional function evaluation. The set of these differences -- i.e., the distance between the objective value of the convex combination and the convex combination of the objective values --  is then used to estimate the probabilities of convexity and linearity. In addition, the arithmetic mean of these differences, as well as the arithmetic mean of the absolute differences are used as features as well.

The 26 \textit{curvature} features measure information on the estimated gradient and Hessian of (a subset of) points from the initial design. More precisely, the lengths' of the gradients, the ratios between the biggest and smallest (absolute) gradient directions and the ratios of the biggest and smallest eigenvalues of the Hessian matrices are aggregated using the minimum, first quartile (= 25\%-quantile), arithmetic mean, median, third quartile (= 75\%-quantile), maximum, standard deviation and number of missing values\footnote{If a point is a local optimum the gradient is zero for all dimensions of a sample point, then the ratio of biggest and smallest gradient obviously can not be computed and therefore results in a missing value (= NA).}. Note that estimating the derivatives also requires additional function evaluations.

The five $y$-\textit{distribution} features compute the kurtosis, skewness and number of peaks of the kernel-based estimation of the density of the initial design's objective values.

For the next group, the initial design is divided into two groups (based on a configurable threshold for the objective values). Afterwards the performances of various classification techniques, which are applied to that binary classification task, are used for computing the 20 \textit{levelset} features. Based on the classifiers and thresholds\footnote{The default classifiers are linear (LDA), quadratic (QDA) and mixture discriminant analysis (MDA) and the default threshold for dividing the data set into two groups are the 10\%-, 25\%- and 50\%-quantile of the objective values.}, \pkg{flacco} uses the \proglang{R}-package \pkg{mlr}~\citep{Mlr2016} to compute the mean misclassification errors (per combination of classifier and threshold). Per default, these error rates are based on a 10-fold crossvalidation, but this can be changed to any other  resampling-strategy that is implemented in \pkg{mlr}. In addition to the ``pure'' misclassification errors, \pkg{flacco} computes the ratio of misclassification errors per threshold and pair of classifiers.

The 15 \textit{local search} features extract information from several local search runs (each of them starting in one of the points from the initial design). Per default, each of the local searches uses a specific quasi-Newton method, namely \texttt{"L-BFGS-B"}~\citep{Byrd1995} as implemented in \proglang{R}'s \texttt{optim}-function, to find a local optimum. The found optima are then clustered (per default via single-linkage hierarchical clustering,~\cite{Jobson2012}) in order to decide whether the found optima belong to the same basins. The resulting total number of found optima (= clusters) is the first feature of this group, whereas the ratio of this number of optima and the conducted local searches form the second one. The next two features measure the ratio of the best and average objective values of the found clusters. As the clusters likely combine multiple local optima, the number of optima per cluster can be used as representation for the basin sizes. These basin sizes are then averaged for the best, all non-best and the worst cluster(s). Finally, the number of required function evaluations across the different local search runs is aggregated using the minimum, first quartile (= 25\%-quantile), arithmetic mean, median, third quartile (= 75\%-quantile), maximum and standard deviation.

The remaining 11 \textit{meta model} features extract information from linear and quadratic models (with and without interactions) that were fitted to the initial design. More precisely, these features compute the model fit (= $R^2_{adj}$) for each of the four aforementioned models, the intercept of the linear model without interactions, the condition of the quadratic model without interactions and the minimum, maximum as well as ratio of maximum and minimum of all (absolute) coefficients -- except for the intercept -- of the linear model without interactions.

\paragraph{Cell mapping features (20 / 3):}
This approach discretizes the continuous decision space using a pre-defined number of blocks (= cells) per input dimension. Then, the interaction of the cells, as well as the observations that are located within a cell, are used to compute the corresponding landscape features~\citep{Kerschke2014}.

\begin{figure*}[!t]
	\begin{minipage}[b]{0.375\textwidth} 
	  \centering
	  \includegraphics[width = 0.9\textwidth]{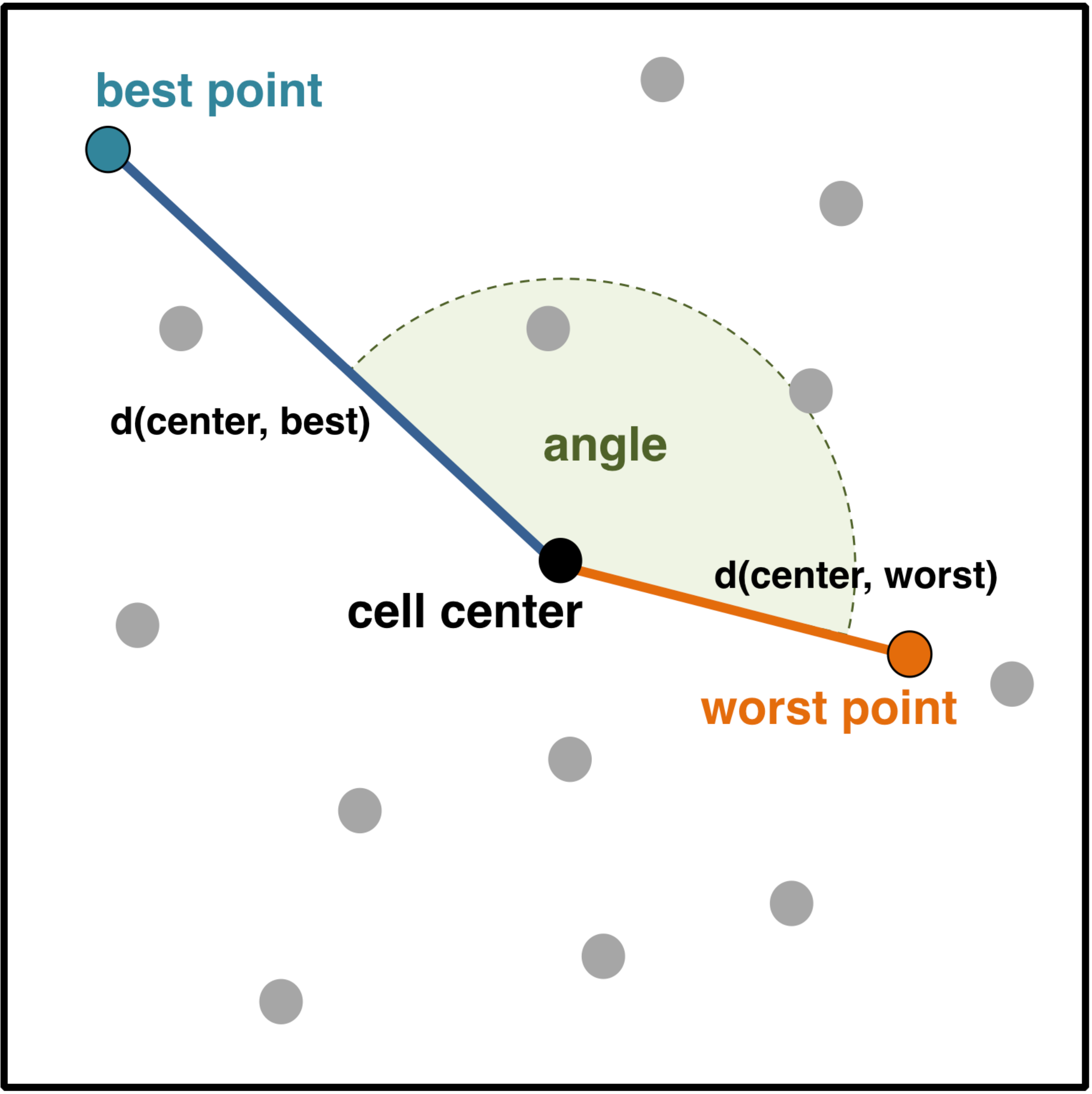}
	  \caption{Overview of the ``ingredients'' computing the \textit{cell mapping angle features}: the location of the best and worst points within a cell.}
	  \label{figure:angle} 
	\end{minipage}
	\hfill
	\begin{minipage}[b]{0.6\textwidth}
	  \centering
	  \includegraphics[width = 0.7425\textwidth]{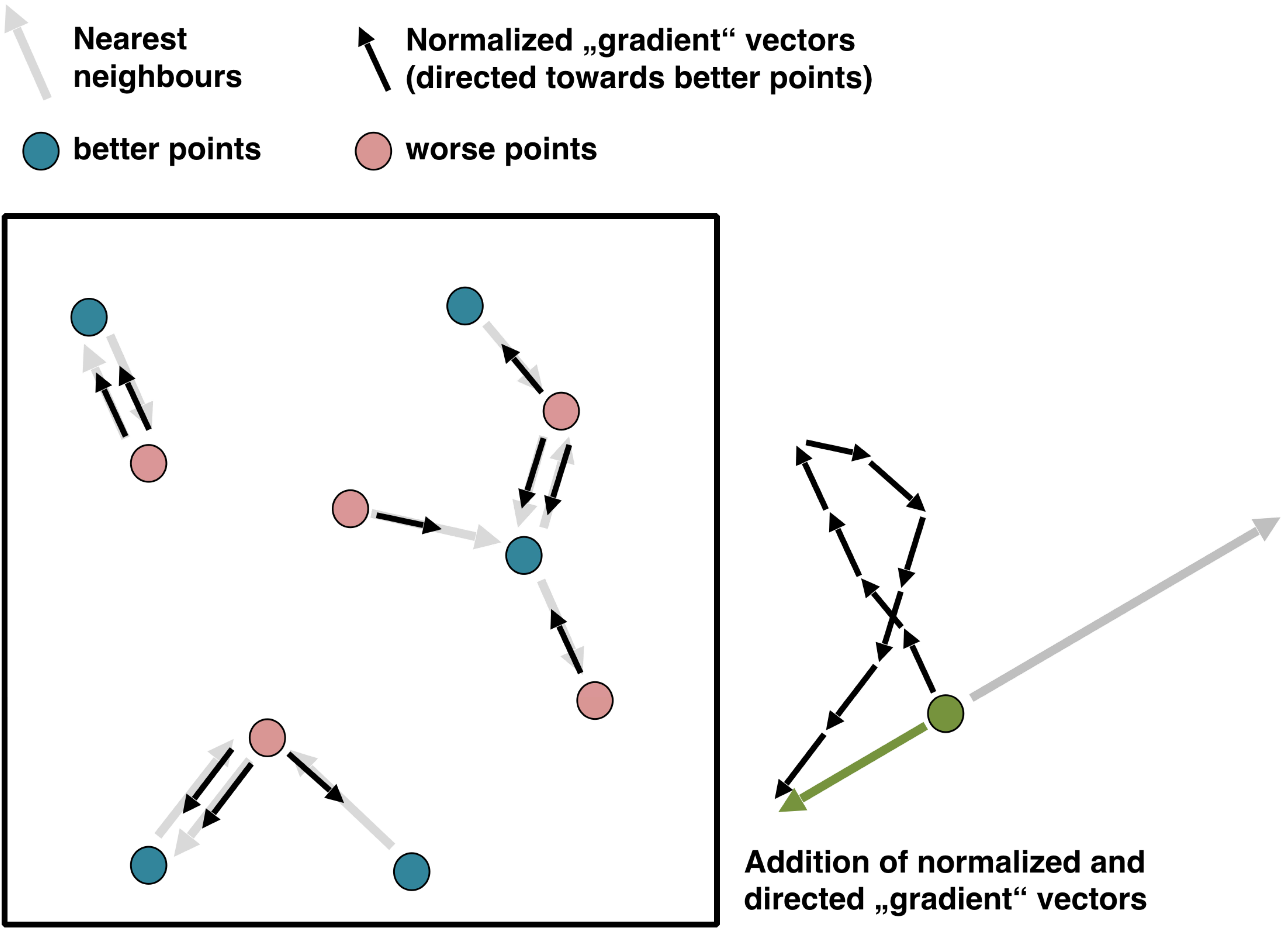}
	\caption{General idea for computing the \textit{cell mapping gradient homogeneity features}: (1) find the nearest neighbours, (2) compute and normalize the gradient vectors, (3) point them towards the better of the two points and (4) sum them up.}
	\label{figure:gradhomo}
	\end{minipage}
\end{figure*}

As shown in Figure~\ref{figure:angle}, the ten \textit{angle} features extract information based on the location of the best and worst (available) observation of a cell~w.r.t.~the corresponding cell center. The distance from cell center to the best and worst point within the cell, as well as the angle between the three points are computed per cell and afterwards aggregated across all cells using the arithmetic mean and standard deviation. The remaining features compute the difference between the best and worst objective value per cell, normalize these differences by the biggest span in objective values within the entire initial design and afterwards aggregate these distances across all cells (using the arithmetic mean and the standard deviation).

The four \textit{gradient homogeneity} features aggregate the cell-wise information on the gradients between each point of a cell and its corresponding nearest neighbor. Figure~\ref{figure:gradhomo} illustrates the idea of this feature set. That is, for each point within a cell, we compute the gradient towards its nearest neighbor, normalize it, point it towards the better one of the two neighboring points and afterwards sum up all the normalized gradients (per cell). Then, the lengths of the resulting vectors are aggregated across all cells using the arithmetic mean and standard deviation.

The remaining six cell mapping features aggregate the (estimated) \textit{convexity} based on representative observations of (all combinations of) three either horizontally, vertically or diagonally successive cells. Each cell is represented by the sample observation that is located closest to the corresponding cell center. We then compare -- for each combination of three neighboring cells -- their objective values: if the objective value of the middle cell is below/above the mean of the objective values of the outer two cells, we have an indication for (soft) convexity/concavity. If the objective value of the middle cell even is the lowest/biggest value of the three neighboring cells, we have an indication for (hard) convexity/concavity. Averaging these values across all combinations of three neighboring cells, results in the estimated ``probabilities'' for (soft/hard) convexity or concavity.

\begin{figure*}[!p]
  \centering
  \includegraphics[width = \textwidth]{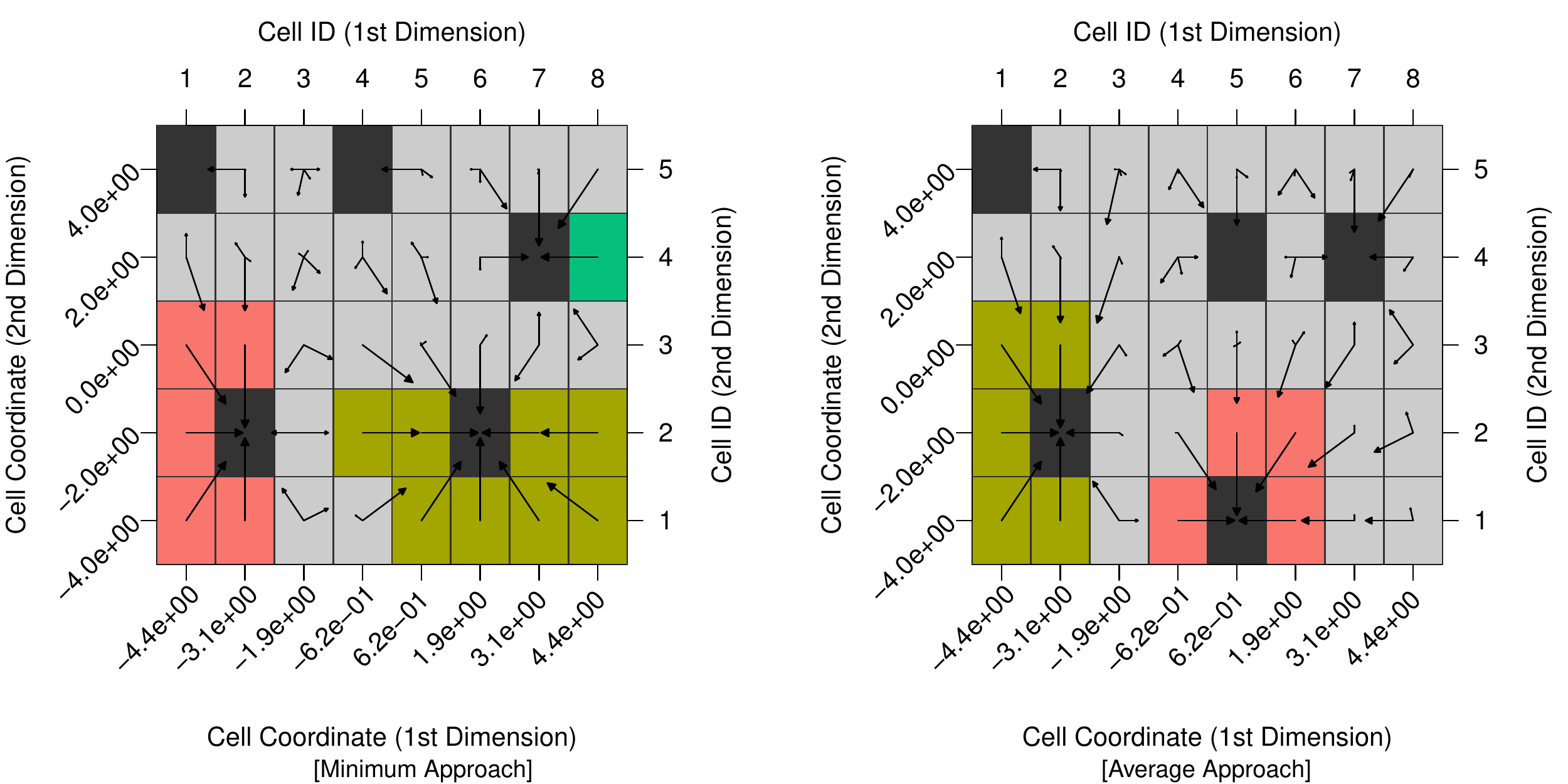}
  \caption{Cell Mappings of a two-dimensional version of \textit{Gallagher's Gaussian 101-me Peaks}~\citep{Hansen2009_Noiseless} function based on a minimum (left) and average (right) approach. The black cells are the attractor cells, i.e.,~potential local optima. Each of those cells might come with a basin of attraction, i.e.,~the colored cells which point towards the attractor. All uncertain cells, i.e.,~cells that are attracted by multiple attractors, are shown in grey. The arrows show in the direction(s) of their attracting cell(s).}
  \label{figure:cm}
  
  \includegraphics[width = \textwidth]{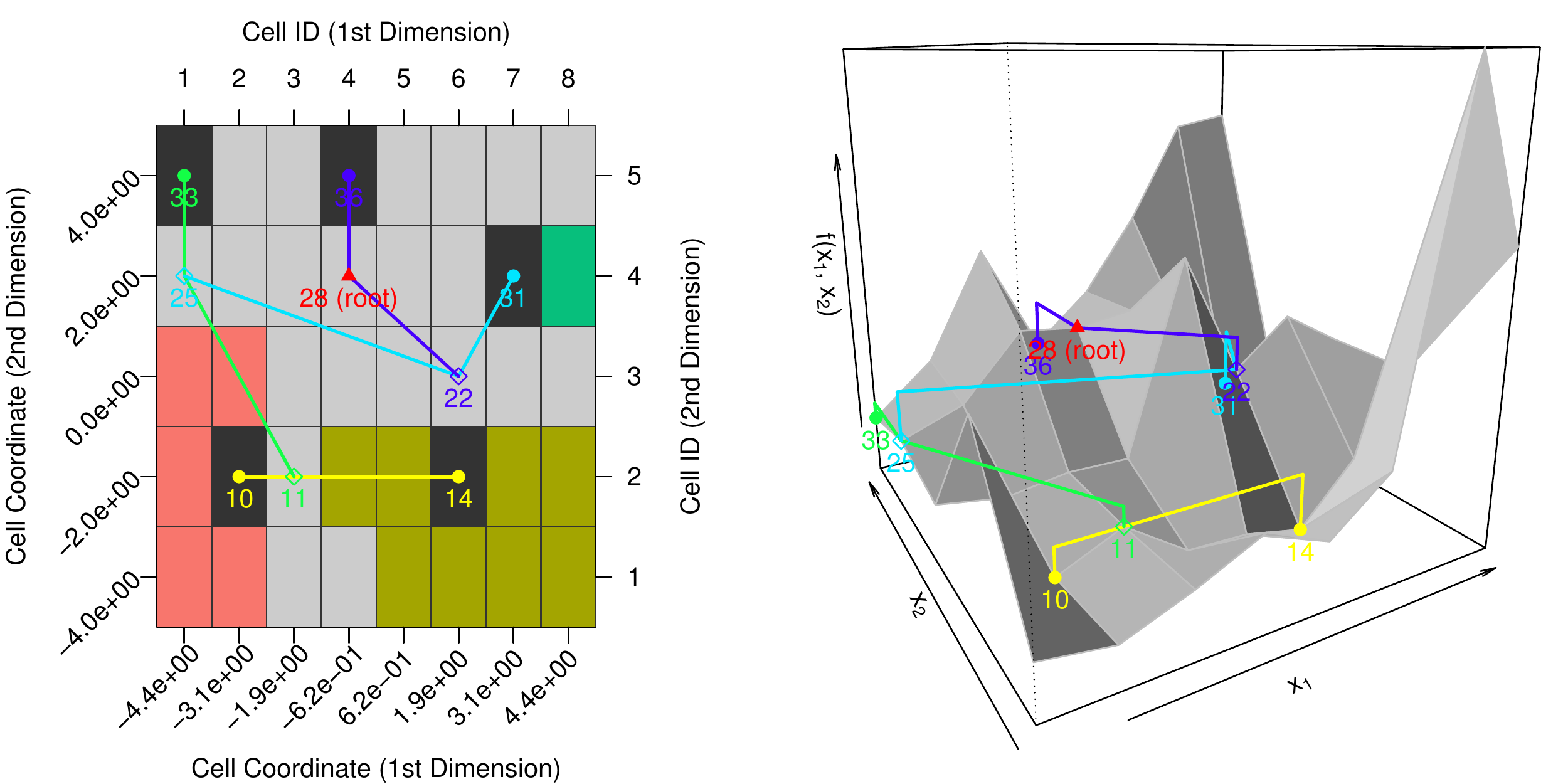}
  \caption{Visualisations of a barrier tree in 2D (left) and 3D (right). Both versions of the tree belong to the same problem -- a two-dimensional version of \textit{Gallagher's Gaussian 101-me Peaks} function based on the minimum approach. The root of the tree is highlighted by a red triangle and the leaves of the tree, which usually lie in an attractor cell, are marked with filled circles. The saddle points, i.e.,~the points where two or more basins of attraction come together, are indicated by empty circles, as well as one incoming and two outgoing branches.}
  \label{figure:bts}
\end{figure*}

\paragraph{Generalized cell mapping features (75 / 1):}
Analogously to the previous group of features, these features are also based on the block-wise discretized decision space. Here, each cell will be represented by exactly one of its observations -- either its best or average value or the one that is located closest to the cell center -- and each of the cells is then considered to be an absorbing Markov chain. That is, for each cell the transition probability for moving from one cell to one of its neighboring cells is computed. Based on the resulting transition probabilities, the cells are categorized into attractor, transient, periodic and uncertain cells. Figure~\ref{figure:cm} shows two exemplary cell mapping plots -- each of them is based on the same feature object, but follows a different representation approach -- which color the cells according to their category: attractor cells are depicted by black boxes, grey cells indicate uncertain cells, i.e., cells that are attracted by multiple attractors, and the remaining cells represent the certain cells, which form the basins of attractions. Furthermore, all non-attractor cells possess arrows that point toward their attractors and their length's represent the attraction probabilities. The different cell types, as well as the accompanying probabilities are the foundation for the 25 GCM-features (per approach):
\begin{enumerate}
\item total number of attractor cells and ratio of cells that are periodic (usually the attractor cells), transient (= non-periodic cells) or uncertain cells,
\item aggregation (minimum, arithmetic mean, median, maximum and standard deviation) of the probabilities for reaching the different basins of attractions,
\item aggregation of the basin sizes when the basins are only formed by the ``certain'' cells,
\item aggregation of the basin sizes when the ``uncertain'' cells also count toward the basins (a cell which points towards multiple attractors contributes to each of them),
\item number and probability of finding the attractor cell with the best objective value.
\end{enumerate}
Further details on the GCM-approach are given in \cite{Kerschke2014}.

\paragraph{Barrier tree features (93 / 1):}
Building on top of the transition probability representation and the three approaches from the generalized cell mapping features, a so-called \emph{barrier tree}~\citep{Flamm2002}, as shown in Figure~\ref{figure:bts}, is constructed (per approach). The local optima of the problem landscape -- i.e.,~the valleys in case of minimization problems -- are represented by the leaves of the tree (indicated by filled circles within Figure~\ref{figure:bts}) and the branching nodes (depicted as non-filled diamonds) represent the ridges of the landscape, i.e.,~the locations where (at least) two neighboring valleys are connected. Based on these trees, the following 31 features can be computed for each of the three cell-representing approaches:
\begin{enumerate}
\item number of leaves (= filled circles) and levels (= branches of different colors), as well as the tree depth, i.e.,~the distance from the root node (red triangle) to the lowest leaf,
\item ratio of the depth of the tree and the number of levels,
\item ratio of the number of levels and the number of non-root nodes,
\item aggregation (minimum, arithmetic mean, median, maximum and standard deviation) of the height differences between a node and its predecessor,
\item aggregation of the average height difference per level,
\item ratio of biggest and smallest basin of attraction (based on the number of cells counting towards the basin) for three different ``definitions'' of a basin: (a) based on the ``certain'' cells, (b) based on the ``uncertain'' cells or (c) based on the ``certain'' cells, plus the ``uncertain'' cells for which the probability towards the respective attractor is the highest,
\item proportion of cells that belong to the basin, which contains the attractor with the best objective value, and cells from the other basins -- aggregated across the different basins,
\item widest range of a basin.
\end{enumerate}

\paragraph{Nearest better clustering features (7 / 1):}
These features extract information based on the comparison of the sets of distances from (a) all observations towards their nearest neighbors and (b) their \textit{nearest better neighbors}\footnote{Here, the ``nearest better neighbor'' is the observation, which is the nearest neighbor among the set of all observations with a better objective value.}~\citep{Kerschke2015}. More precisely, these features measure the ratios of the standard deviations and the arithmetic means between the two sets, the correlation between the distances of the nearest neighbors and nearest better neighbors, the coefficient of variation of the distance ratios and the correlation between the fitness value, and the count of observations to whom the current observation is the nearest better neighbour, i.e.,~the so-called ``indegree''.

\paragraph{Dispersion features (18 / 1):}
The \textit{dispersion features} by~\cite{Lunacek2006} compare the dispersion among observations within the initial design and among a subset of these points. The subsets are created based on predefined thresholds, whose default values are the 2\%-, 5\%-, 10\%- and 25\%-quantile of the objective values. For each of these threshold values, we compute the arithmetic mean and median of all distances among the points of the subset, and then compare it -- using the difference and ratio -- to the mean or median, respectively, of the distances among all points from the initial design.

\paragraph{Information content features (7 / 1):}
The \textit{Information Content of Fitness Sequences (ICoFiS)} approach~\citep{Munoz2015} quantifies the so-called \textit{information content} of a continuous landscape, i.e.,~smoothness, ruggedness, or neutrality. While similar methods already exist for the information content of discrete landscapes, this approach provides an adaptation to continuous landscapes that for instance accounts for variable step sizes in random walk sampling. This approach is based on a symbol sequence $\Phi = \{\phi_1, \ldots, \phi_{n-1}\}$, with

\[\phi_i := \begin{cases} \bar{1} & \text{, if } \phantom{||||}\frac{y_{i + 1} - y_i}{||\mathbf{x}_{i + 1} - \mathbf{x}_i||}\phantom{|||} < - \varepsilon \\ 0 & \text{, if } \phantom{a}\left|\frac{y_{i + 1} - y_i}{||\mathbf{x}_{i + 1} - \mathbf{x}_i||}\right|\phantom{a} \le \phantom{-}\varepsilon \\ 1 & \text{, if } \phantom{||||}\frac{y_{i + 1} - y_i}{||\mathbf{x}_{i + 1} - \mathbf{x}_i||}\phantom{|||} > \phantom{-}\varepsilon \end{cases}.\]

This sequence is derived from the objective values $y_1, \ldots, y_n$ belonging to the $n$ points $\mathbf{x}_1, \ldots, \mathbf{x}_n$ of a random walk across (the initial design of) the landscape and depends on the \textit{information sensitivity} parameter $\varepsilon > 0$.

This symbol sequence $\Phi$ is aggregated by the \textit{information content} $H(\varepsilon) := \sum_{i \ne j} p_{ij} \cdot \log_{6} p_{ij}$, where $p_{ij}$ is the probability of having the ``block'' $\phi_i\phi_j$, with $\phi_i, \phi_j \in \{\bar{1}, 0, 1\}$, within the sequence. Note that the base of the logarithm was set to six as this equals the number of possible blocks $\phi_i\phi_j$ for which $\phi_i \ne \phi_j$, i.e., $\phi_i\phi_j \in \{\bar{1}0, 0\bar{1}, \bar{1}1, 1\bar{1}, 10, 01\}$~\citep{Munoz2015}.

Another aggregation of the information is the so-called  \textit{partial information content} $M(\varepsilon) := |\Phi^{'}| / (n - 1)$, where $\Phi^{'}$ is the symbol sequence of alternating $1$'s and $\bar{1}$'s, which is derived from $\Phi$ by removing all zeros and repeated symbols. \cite{Munoz2015} then suggest to use these two characteristics for computing the following five features:
\begin{enumerate}
\item \textit{maximum information content} $H_{max} := \max_{\varepsilon} \{H(\varepsilon)\}$,
\item \textit{settling sensitivity} $\varepsilon_{s} := \log_{10} \left( \min_{\varepsilon} \{ \varepsilon: H(\varepsilon) < s \} \right)$, with the default of $s$ being $0.05$ as suggested by \cite{Munoz2015},
\item $\varepsilon_{max} := \arg\max_{\varepsilon} \{H(\varepsilon)\}$,
\item \textit{initial partial information} $M_{0} := M(\varepsilon = 0)$,
\item \textit{ratio of partial information sensitivity} $\varepsilon_{r} := \log_{10} \left( \max_{\varepsilon} \{ \varepsilon: M(\varepsilon) > r \cdot M_{0} \} \right)$, with the default of $r$ being $0.5$ as suggested by \cite{Munoz2015}.
\end{enumerate}

\begin{figure*}[!t]
	  \centering
	  \includegraphics[width = 0.75\textwidth]{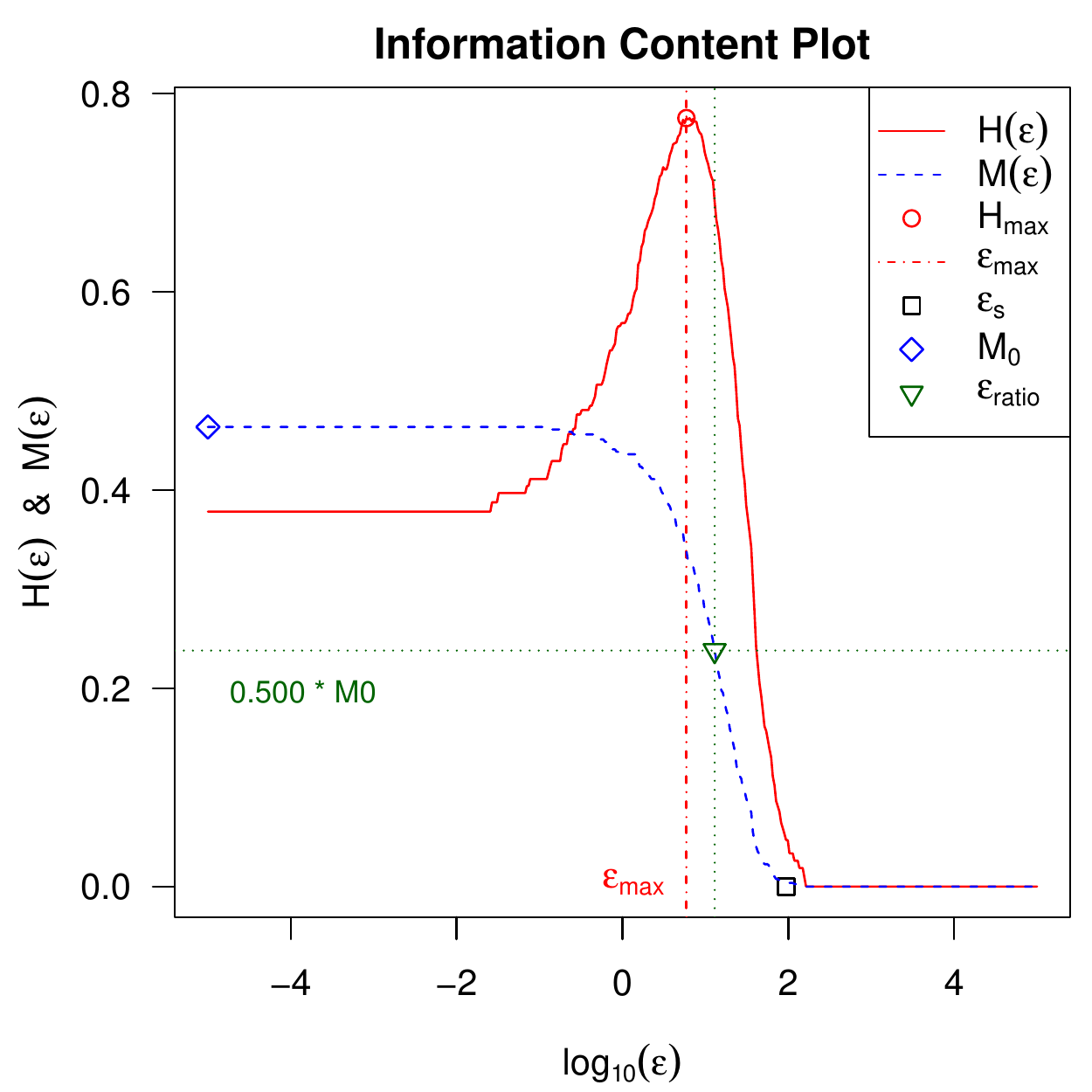}
	  \caption{\textit{Information Content Plot} of a two-dimensional version of \textit{Gallagher's Gaussian 101-me Peaks} function. This approach analyzes the behavior of the tour along the points of the problem's initial design.}
	  \label{figure:gallaghers_ic2} 
\end{figure*}

The various characteristics and features described above can also be visualized within an \textit{Information Content Plot} as exemplarily shown in Figure~\ref{figure:gallaghers_ic2}.

\paragraph{Further miscellaneous features (40 / 3):}
The remaining three feature sets provided within \pkg{flacco} are based on some very simple ideas. Note that there exists no (further) detailed description for these three feature sets as -- according to the best knowledge of the author -- none of them has been used for any (published) experiments. However, from our perspective they are worth to be included in future studies.

The simplest and quickest computable group of features are the 16 \textit{basic} features, which provide rather obvious information of the initial design: (a) the number of observations and input variables, (b) the minimum and maximum of the lower and upper boundaries, objective values and number of blocks per dimension, (c) the number of filled cells and the total number of cells, and (d) a binary flag stating whether the objective function should be minimized.

The 14 \textit{linear model} features fit a linear model -- with the objective variable being the depending variable and all the remaining variables as explaining variables -- within each cell of the feature object and aggregate the coefficient vectors across all the cells. That is, we compute (a) the length of the average coefficient vector, (b) the correlation of the coefficient vectors, (c) the ratio of maximum and minimum, as well as the arithmetic mean of the standard deviation of the (non-intercept) coefficients, (d) the previous features based on coefficient vectors that were a priori normalized per cell, (e) the arithmetic mean and standard deviation of the lengths of the coefficient vectors (across the cells), and (f) the arithmetic mean and standard deviation of the ratio of biggest and smallest (non-intercept) coefficients per cell.

The remaining ten features extract information from a \textit{principal component} analysis, which is performed on the initial design -- both, including and excluding the objective values -- and that are based on the covariance, as well as correlation matrix, respectively. For each of these four combinations, the features measure the proportion of principal components that are needed to explain a pre-defined percentage (default: 0.9) of the data's variance, as well as the percentage of variation that is explained by the first principal component.

Further information on all the implemented features, especially more detailed information on each of its configurable parameters is given within the documentation of the package. Note that -- in contrast to the original implementation of most of the feature sets -- the majority of parameters, e.g.,~the classifiers used by the level set features, or the ratio for the partial information sensitivity within the information content features, are configurable. While for reasons of conveniency, the default values are set according to their original publications, they can easily be changed using the \texttt{control} argument within the functions \texttt{calculateFeatureSet} or \texttt{calculateFeatures}, respectively. An example for this is given at the end of Section~\ref{sec:examples}.

Note that each feature set can be computed for unconstrained, as well as box-constrained optimization problems. Upon creation of the feature object, \pkg{R} asserts that each point from the initial sample \texttt{X} lies within the defined \texttt{lower} and \texttt{upper} boundaries. Hence, if one intends to consider unconstrained problems, one should set the boundaries to \texttt{-Inf} or \texttt{Inf}, respectively. When dealing with box-constrained problems, one has to take a look at the feature sets themselves. Given that 14 out of the 17 feature sets do not perform additional function evaluations at all, none of them can violate any of the constraints. This also holds for the remaining three feature sets. The \textit{convexity} features always evaluate (additional) points that are located in the center of two already existing (and thereby feasible) points. Consequently, each of them has to be located within the box-constraints as well. In case of the \textit{local search} features, the box-constraints are also respected as long as one uses the default local search algorithm (L-BFGS-B), which was designed to optimize bound-constrained optimization problems. Finally, the \textit{curvature} features do not violate the constraints either as the internal functions, which are used for estimating the gradient and Hessian of the function, were adapted in such a way that they always estimate them by evaluating points within the box-constraints -- even in case the points are located close to or even exactly on the border of the box.

As described within the previous two sections, numerous researchers have already developed feature sets and many of them also shared their source code, enabling the creation of this wide collection of landscape features in the first place. By developing \pkg{flacco} on a publicly accessible platform\footnote{The development version is available on GitHub: \url{https://github.com/kerschke/flacco}}, other \proglang{R}-users may easily contribute to the package by providing further feature sets. Aside from using the most recent (development) version of the package, one can also use the stable release of \pkg{flacco} provided on CRAN\footnote{The stable release is published on CRAN: \url{https://cran.r-project.org/package=flacco}}. Note that the package repository on GitHub also provides a link to the corresponding online tutorial\footnote{Link to the package's tutorial: \url{http://kerschke.github.io/flacco-tutorial/}}, the platform-independent web-application\footnote{Link to the package's GUI: \url{https://flacco.shinyapps.io/flacco/}} (further details are given in Section~\ref{sec:gui}) and an issue tracker, where one could report bugs, provide feedback or suggest the integration of further feature sets.


\section{Exemplary feature computation}\label{sec:examples}

\begin{figure*}[!b]
	\begin{minipage}[t]{0.495\textwidth} 
	  \centering
	  \includegraphics[width = \textwidth]{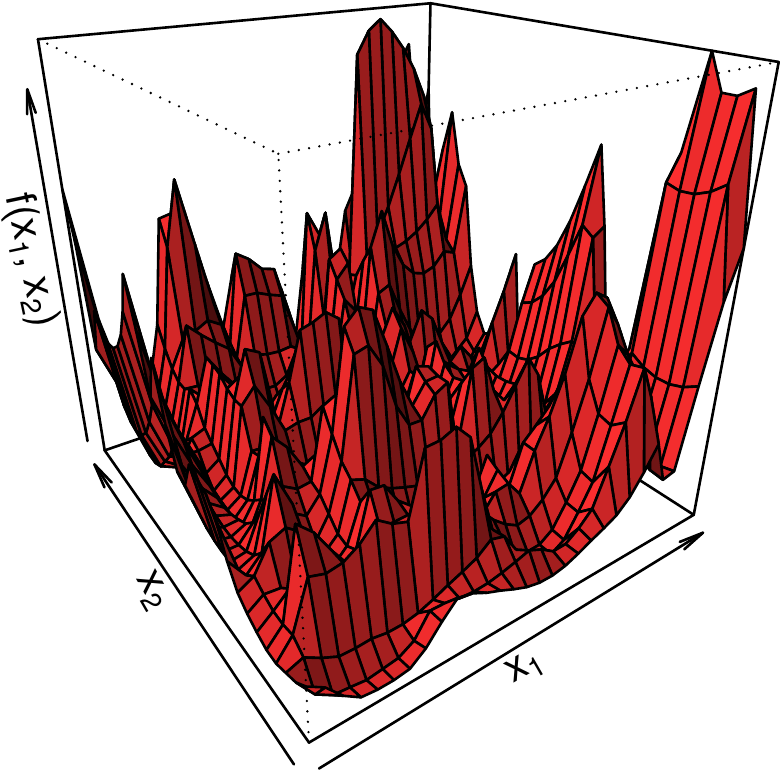}
	  \caption{3D-Perspective plot of \textit{Gallagher's Gaussian 101-me Peaks} function.}
	  \label{figure:fig4} 
	\end{minipage}
	\hfill
	\begin{minipage}[t]{0.495\textwidth}
	  \centering
	  \includegraphics[width = \textwidth]{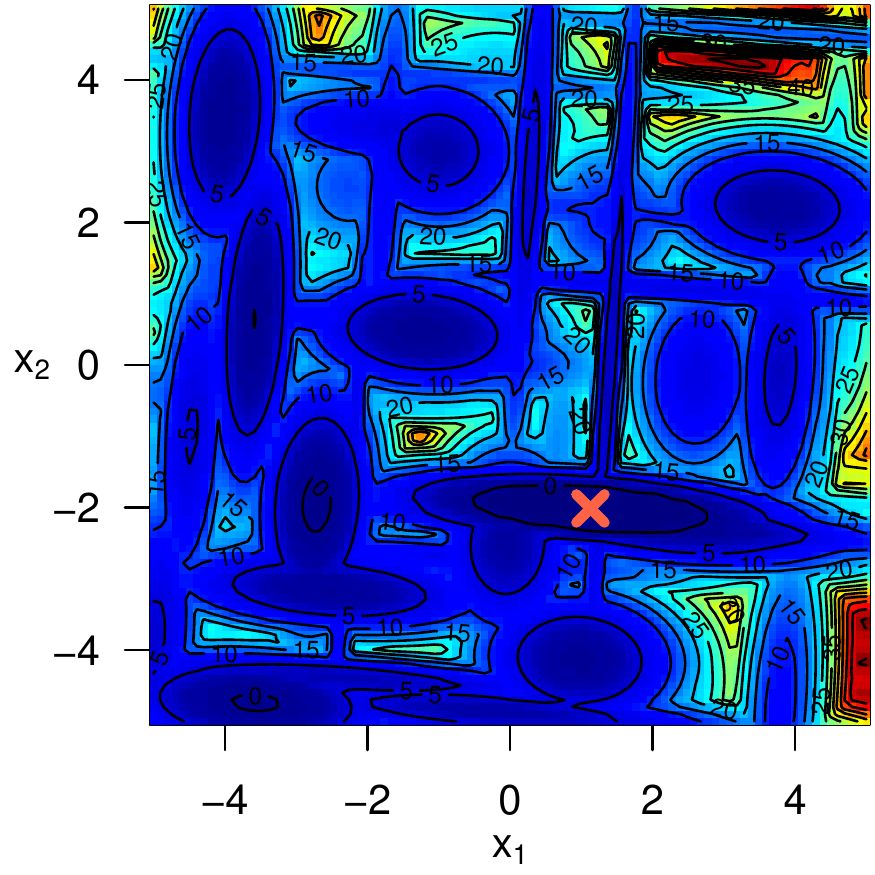}
	  \caption{Contour plot of \textit{Gallagher's Gaussian 101-me Peaks} function. The red cross marks the global optimum and the coloring of the plot represents the objective values.}
	  \label{figure:fig5}
	\end{minipage}
\end{figure*}

In the following, the usage of \pkg{flacco} will exemplarily be presented on a well-known Black-Box optimization problem, namely \textit{Gallagher's Gaussian 101-me Peaks}~\citep{Hansen2009_Noiseless}. This is the 21st out of the 24 artificially designed, continuous single-objective optimization problems from the \textit{Black-Box Optimization Benchmark} (BBOB,~\cite{Hansen2010}). Within this benchmark, it belongs to a group of five multimodal problem instances with a weak global structure. Each of the 24 function classes from BBOB can be seen as a specific problem generator, whose problems are identical up to rotation, shifts and scaling. Therefore the exact instance has to be defined by a function ID (\texttt{fid}) and an instance ID (\texttt{iid}). For this exemplary presentation, we will use the second instance and generate it using \texttt{makeBBOBFunction} from the \proglang{R}-package~\pkg{smoof}~\citep{Smoof2016}. The resulting landscape is depicted as a three-dimensional perspective plot in Figure~\ref{figure:fig4} and as a contour plot in Figure~\ref{figure:fig5}.

In a first step, one needs to install the package along with all its dependencies, load it into the workspace and afterwards generate the input data -- i.e.,~a (hopefully) representative sample \texttt{X} of observations from the decision space along with their corresponding objective values \texttt{y}.

\begin{knitrout}
\definecolor{shadecolor}{rgb}{0.969, 0.969, 0.969}\color{fgcolor}\begin{kframe}
\begin{alltt}
\hlstd{R> }\hlkwd{install.packages}\hlstd{(}\hlstr{"flacco"}\hlstd{,} \hlkwc{dependencies} \hlstd{=} \hlnum{TRUE}\hlstd{)}
\hlstd{R> }\hlkwd{library}\hlstd{(}\hlstr{"flacco"}\hlstd{)}
\hlstd{R> }\hlstd{f} \hlkwb{=} \hlstd{smoof}\hlopt{::}\hlkwd{makeBBOBFunction}\hlstd{(}\hlkwc{dimension} \hlstd{=} \hlnum{2}\hlstd{,} \hlkwc{fid} \hlstd{=} \hlnum{21}\hlstd{,} \hlkwc{iid} \hlstd{=} \hlnum{2}\hlstd{)}
\hlstd{R> }\hlstd{ctrl} \hlkwb{=} \hlkwd{list}\hlstd{(}\hlkwc{init_sample.lower} \hlstd{=} \hlopt{-}\hlnum{5}\hlstd{,} \hlkwc{init_sample.upper} \hlstd{=} \hlnum{5}\hlstd{)}
\hlstd{R> }\hlstd{X} \hlkwb{=} \hlkwd{createInitialSample}\hlstd{(}\hlkwc{n.obs} \hlstd{=} \hlnum{800}\hlstd{,} \hlkwc{dim} \hlstd{=} \hlnum{2}\hlstd{,} \hlkwc{control} \hlstd{= ctrl)}
\hlstd{R> }\hlstd{y} \hlkwb{=} \hlkwd{apply}\hlstd{(X,} \hlnum{1}\hlstd{, f)}
\end{alltt}
\end{kframe}
\end{knitrout}

The code above would also work without explicitly specifying the lower and upper boundaries (within \texttt{ctrl}), but in that case, the decision sample would automatically be generated within $[0, 1] \times [0, 1]$ rather than within the (for BBOB) common box-constraints of $[-5, 5] \times [-5, 5]$. In a next step, this input data is used (eventually in combination with further parameters such as the number of blocks per dimension) to create the feature object as already schematized in Figure~\ref{figure:featobject}. The following \proglang{R}-code shows how this could be achieved.

\begin{knitrout}
\definecolor{shadecolor}{rgb}{0.969, 0.969, 0.969}\color{fgcolor}\begin{kframe}
\begin{alltt}
\hlstd{R> }\hlstd{feat.object} \hlkwb{=} \hlkwd{createFeatureObject}\hlstd{(}\hlkwc{X} \hlstd{= X,} \hlkwc{y} \hlstd{= y,} \hlkwc{fun} \hlstd{= f,} \hlkwc{blocks} \hlstd{=} \hlkwd{c}\hlstd{(}\hlnum{8}\hlstd{,} \hlnum{5}\hlstd{))}
\end{alltt}
\end{kframe}
\end{knitrout}

As stated earlier, a feature object is the fundamental object for the majority of computations performed by \pkg{flacco} and thus, it obviously has to store quite a lot of information -- such as the number of observations and (input) variables, the names and boundaries of the variables, the amount of empty and non-empty cells, etc. Parts of that information can easily be printed to the console by calling \texttt{print} on the feature object.\footnote{Note that the shown numbers, especially the ones for the number of observations per cell, might be different on your machine, as the initial sample is drawn randomly.}

\begin{knitrout}
\definecolor{shadecolor}{rgb}{0.969, 0.969, 0.969}\color{fgcolor}\begin{kframe}
\begin{alltt}
\hlstd{R> }\hlkwd{print}\hlstd{(feat.object)}
\end{alltt}
\begin{verbatim}
Feature Object:
- Number of Observations: 800
- Number of Variables: 2
- Lower Boundaries: -5.00e+00, -5.00e+00
- Upper Boundaries: 5.00e+00, 5.00e+00
- Name of Variables: x1, x2
- Optimization Problem: minimize y
- Function to be Optimized: smoof-function (BBOB_2_21_2)
- Number of Cells per Dimension: 8, 5
- Size of Cells per Dimension: 1.25, 2.00
- Number of Cells:
  - total: 40
  - non-empty: 40 (100.00%)
  - empty: 0 (0.00%)
- Average Number of Observations per Cell:
  - total: 20.00
  - non-empty: 20.00
\end{verbatim}
\end{kframe}
\end{knitrout}

Given the feature object, one can easily calculate any of the 17 feature sets that were introduced in Section~\ref{sec:ela}. Specific feature sets can for instance -- exemplarily shown for the \textit{angle}, \textit{dispersion} and \textit{nearest better clustering} features -- be computed as shown in the code below. 

\begin{knitrout}
\definecolor{shadecolor}{rgb}{0.969, 0.969, 0.969}\color{fgcolor}\begin{kframe}
\begin{alltt}
\hlstd{R> }\hlstd{angle.features} \hlkwb{=} \hlkwd{calculateFeatureSet}\hlstd{(}
\hlstd{+ }  \hlkwc{feat.object} \hlstd{= feat.object,} \hlkwc{set} \hlstd{=} \hlstr{"cm_angle"}\hlstd{)}
\hlstd{R> }\hlstd{dispersion.features} \hlkwb{=} \hlkwd{calculateFeatureSet}\hlstd{(}
\hlstd{+ }  \hlkwc{feat.object} \hlstd{= feat.object,} \hlkwc{set} \hlstd{=} \hlstr{"disp"}\hlstd{)}
\hlstd{R> }\hlstd{nbc.features} \hlkwb{=} \hlkwd{calculateFeatureSet}\hlstd{(}
\hlstd{+ }  \hlkwc{feat.object} \hlstd{= feat.object,} \hlkwc{set} \hlstd{=} \hlstr{"nbc"}\hlstd{)}
\end{alltt}
\end{kframe}
\end{knitrout}

Alternatively, one can compute all of the more than 300 features simultaneously via:

\begin{knitrout}
\definecolor{shadecolor}{rgb}{0.969, 0.969, 0.969}\color{fgcolor}\begin{kframe}
\begin{alltt}
\hlstd{R> }\hlstd{all.features} \hlkwb{=} \hlkwd{calculateFeatures}\hlstd{(feat.object)}
\end{alltt}
\end{kframe}
\end{knitrout}

Each of these calculations results in a list of (mostly numeric, i.e.,~real-valued) features. In order to avoid errors due to the possibility of similar feature names between different feature sets, each feature inherits the abbreviation of its feature set's name as a prefix, e.g.,~\texttt{cm\_angle} for the \textit{cell mapping angle} features or \texttt{disp} for the \textit{dispersion} features. Using the results from the previous example, the output for the calculation of the seven \textit{nearest better clustering} features could look like this:

\begin{knitrout}
\definecolor{shadecolor}{rgb}{0.969, 0.969, 0.969}\color{fgcolor}\begin{kframe}
\begin{alltt}
\hlstd{R> }\hlkwd{str}\hlstd{(nbc.features)}
\end{alltt}
\begin{verbatim}
List of 7
 $ nbc.nn_nb.sd_ratio      : num 0.303
 $ nbc.nn_nb.mean_ratio    : num 0.605
 $ nbc.nn_nb.cor           : num 0.271
 $ nbc.dist_ratio.coeff_var: num 0.383
 $ nbc.nb_fitness.cor      : num -0.364
 $ nbc.costs_fun_evals     : int 0
 $ nbc.costs_runtime       : num 0.061
\end{verbatim}
\end{kframe}
\end{knitrout}

At this point, the authors would like to emphasize that \textbf{one should not try to interpret the numerical feature values} on their own as the majority of them simply are not intuitively understandable. Instead, these numbers should rather be used to distinguish between different problems -- usually in an automated fashion, e.g., by means of a machine learning algorithm.

Also, it is important to note that features, which belong to the same feature set, are based on the same idea and only differ in the way they aggregate those ideas. For instance, the \textit{nearest better clustering} features are based on the same distance sets -- the distances to their nearest neighbors and nearest better neighbors. However, the features differ in the way they aggregate those distance sets, e.g.,~by computing the ratio of their standard deviations or the ratio of their arithmetic means. As the underlying approaches are the bottleneck of each feature set, computing single features on their own would not be very beneficial and therefore has not been implemented in this package. Therefore, given the fact that multiple features share a common idea, which is usually the expensive part of the feature computation -- for this group of features it would be the computation of the two distance sets -- they will be computed together as an entire feature set instead of computing each feature on its own.

\begin{figure*}[!t]
  \centering
  \includegraphics[width = 0.75\textwidth]{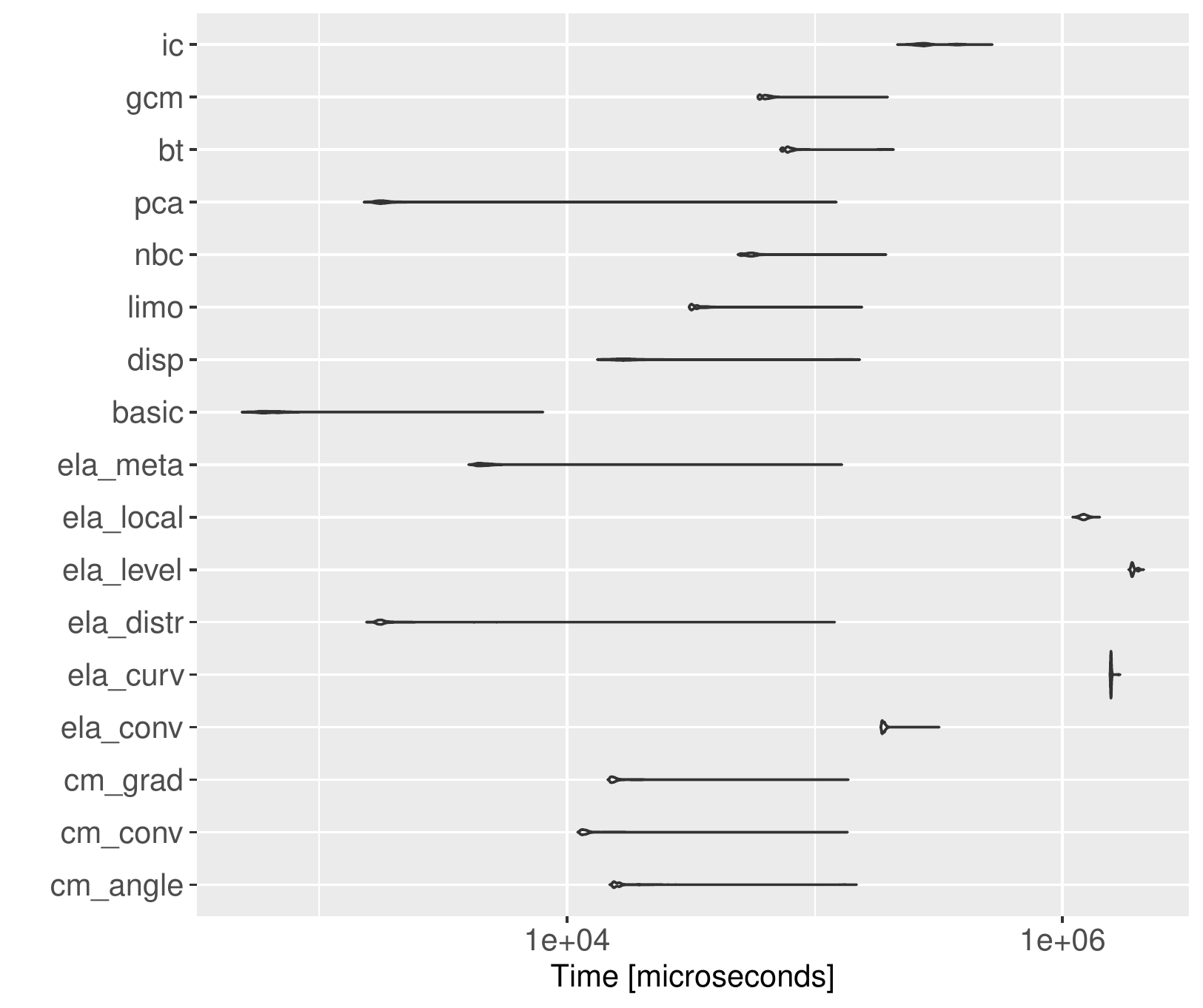}
  \caption{Visualization of a microbenchmark, measuring the logarithmic runtimes (in microseconds) of each feature set based on 1,000 replications on a two-dimensional version of the \textit{Gallagher's Gaussian 101-me Peaks} function.}
  \label{figure:benchmark}
\end{figure*}

Figure \ref{figure:benchmark} shows the results of a so-called ``microbenchmark'' \citep{Microbenchmark2014} on the chosen BBOB instance. Such a benchmark allows the comparison of the (logarithmic) runtimes across the different feature sets. As one can see, the time for calculating a specific feature set heavily depends on the chosen feature set. Although the majority of feature sets can be computed in less than half a second (the \textit{basic} features even in less than a millisecond), three feature sets -- namely \textit{curvature}, \textit{level set} and \textit{local search} -- usually needed between one and two seconds. And although two seconds might not sound that bad, it can have a huge impact on the required resources when computing all feature sets across thousands or even millions of instances. The amount of required time might even increase when considering the fact that some of the features are stochastic. In such a case it is strongly recommended to calculate each of those feature sets multiple times (on the same problem instance) in order to capture the variance of these features.

Note that the current definition of the feature object also contains the actual \textit{Gallagher's Gaussian 101-me Peaks} function itself -- it was added to the feature object by setting the parameter \texttt{fun = f} within the function \texttt{createFeatureObject}. Without that information it would have been impossible to calculate feature sets, which require additional function evaluations, i.e.,~the \textit{convexity}, \textit{curvature} and \textit{local search} features of the classical ELA features. Similarly, the \textit{barrier tree}, \textit{cell mapping} and \textit{general cell mapping} features strongly depend on the value of the \texttt{blocks} argument, which defines the number of blocks per dimension when representing the decision space as a grid of cells.\footnote{The \textit{barrier tree} features can only be computed if the total number of cells is at least two and the \textit{cell mapping convexity} features require at least three blocks per dimension.}

As mentioned in Section~\ref{sec:ela}, the majority of feature sets uses some parameters, all of them having certain default settings. However, as different landscapes / optimization problems might require different configurations of these parameters, the package allows its users to change the default parameters by setting their parameter configuration within the \texttt{control} argument of the functions \texttt{calculateFeatures} or \texttt{calculateFeatureSet}, respectively. Similarly to the naming convention of the features themselves, the names of parameters use the abbreviation of the corresponding feature set as a prefix. This way, one can store all the preferred configurations within a single control argument and does not have to deal with a separate control argument for each feature set. In order to illustrate the usage of the control argument, let's assume one wants to solve the following two tasks:
\begin{enumerate}
\item Calculate the \textit{dispersion} features for a problem, where the Manhattan distance (also known as city-block distance or L$_1$-norm) is more reasonable than the (default) Euclidean distance.
\item Calculate the \textit{General Cell Mapping} features for two instead of the usual three GCM approaches (i.e.,~\texttt{"min"} = best value per cell, \texttt{"mean"} = average value per cell and \texttt{"near"} = closest value to a cell center).
\end{enumerate}

Due to the availability of a shared control parameter, these adaptations can easily be performed with a minimum of additional code:

\begin{knitrout}
\definecolor{shadecolor}{rgb}{0.969, 0.969, 0.969}\color{fgcolor}\begin{kframe}
\begin{alltt}
\hlstd{R> }\hlstd{ctrl} \hlkwb{=} \hlkwd{list}\hlstd{(}
\hlstd{+ }  \hlkwc{disp.dist_method} \hlstd{=} \hlstr{"manhattan"}\hlstd{,}
\hlstd{+ }  \hlkwc{gcm.approaches} \hlstd{=} \hlkwd{c}\hlstd{(}\hlstr{"min"}\hlstd{,} \hlstr{"near"}\hlstd{)}
\hlstd{+ }\hlstd{)}
\hlstd{R> }\hlstd{dispersion.features} \hlkwb{=} \hlkwd{calculateFeatureSet}\hlstd{(}
\hlstd{+ }  \hlkwc{feat.object} \hlstd{= feat.object,} \hlkwc{set} \hlstd{=} \hlstr{"disp"}\hlstd{,} \hlkwc{control} \hlstd{= ctrl)}
\hlstd{R> }\hlstd{gcm.features} \hlkwb{=} \hlkwd{calculateFeatureSet}\hlstd{(}
\hlstd{+ }  \hlkwc{feat.object} \hlstd{= feat.object,} \hlkwc{set} \hlstd{=} \hlstr{"gcm"}\hlstd{,} \hlkwc{control} \hlstd{= ctrl)}
\end{alltt}
\end{kframe}
\end{knitrout}

This shared control parameter also works when computing all features simultaneously:

\begin{knitrout}
\definecolor{shadecolor}{rgb}{0.969, 0.969, 0.969}\color{fgcolor}\begin{kframe}
\begin{alltt}
\hlstd{R> }\hlstd{all.features} \hlkwb{=} \hlkwd{calculateFeatures}\hlstd{(feat.object,} \hlkwc{control} \hlstd{= ctrl)}
\end{alltt}
\end{kframe}
\end{knitrout}

A detailed overview of all the configurable parameters is available within the documentation of the previous functions and can for instance be accessed via the command \texttt{?calculateFeatures}.


\section{Visualisation techniques}\label{sec:visual}

In addition to the features themselves, \pkg{flacco} provides a handful of visualization techniques allowing the user to get a better understanding of the features. The function \texttt{plotCellMapping} produces a plot of the discretized grid of cells, representing the problem's landscape when using the cell mapping approach. Figure \ref{figure:cm}, which was already presented in Section~\ref{sec:ela}, shows two of these cell mapping plots -- the left one represents each cell by the smallest fitness value of its respective points, and the right plot represents each cell by the average of the fitness values of the respective cell's sample points. In each of those plots, the \textit{attractor cells}, i.e.,~cells containing the sample's local optima, are filled black. The \textit{uncertain cells}, i.e.,~cells that transition to multiple attractors, are colored grey, whereas the \textit{certain cells} are colored according to their \textit{basin of attraction}, i.e.,~all cells which transition towards the same attractor (and no other one) share the same color. Additionally, the figure highlights the attracting cells of each cell via arrows that point towards the corresponding attractor(s). Note that the length of the arrows is proportional to the transition probabilities.

Given the feature object from the example of the previous section, i.e.,~the one that is based on \textit{Gallagher's Gaussian 101-me Peaks} function, the two cell mapping plots can be produced with the following code:

\begin{knitrout}
\definecolor{shadecolor}{rgb}{0.969, 0.969, 0.969}\color{fgcolor}\begin{kframe}
\begin{alltt}
\hlstd{R> }\hlkwd{plotCellMapping}\hlstd{(feat.object,}
\hlstd{+ }  \hlkwc{control} \hlstd{=} \hlkwd{list}\hlstd{(}\hlkwc{gcm.approach} \hlstd{=} \hlstr{"min"}\hlstd{))}
\hlstd{R> }\hlkwd{plotCellMapping}\hlstd{(feat.object,}
\hlstd{+ }  \hlkwc{control} \hlstd{=} \hlkwd{list}\hlstd{(}\hlkwc{gcm.approach} \hlstd{=} \hlstr{"mean"}\hlstd{))}
\end{alltt}
\end{kframe}
\end{knitrout}

Analogously to the cell mapping plots, one can visualize the barrier trees of a problem. They can either be plotted in 2D as a layer on top of a cell mapping (per default without any arrows) or in 3D as a layer on top of a perspective~/~surface plot of the discretized decision space. Using the given feature object, the code for creating the plots from Figure \ref{figure:bts} (also already shown in Section~\ref{sec:ela}) would look like this:

\begin{knitrout}
\definecolor{shadecolor}{rgb}{0.969, 0.969, 0.969}\color{fgcolor}\begin{kframe}
\begin{alltt}
\hlstd{R> }\hlkwd{plotBarrierTree2D}\hlstd{(feat.object)}
\hlstd{R> }\hlkwd{plotBarrierTree3D}\hlstd{(feat.object)}
\end{alltt}
\end{kframe}
\end{knitrout}

The representation of the tree itself is in both cases very similar. Each of the trees begins with its root, depicted by a red triangle. Starting from that point, there will usually be two or more branches, pointing either towards a saddle point (depicted by a non-filled diamond) -- a node that connects multiple basins of attraction -- or towards a leaf (filled circles) of the tree, indicating a local optimum. Branches belonging to the same level (i.e.,~having the same number of predecessors on their path up to the root) of the tree are colored identically. Note that all of the aforementioned points, i.e.,~root, saddle points and leaves, belong to distinct cells. The corresponding (unique) cell IDs are given by the numbers next to the points.

The last plot, which is directly related to the visualization of a feature set, is the so-called \textit{information content plot}, which was shown in Figure~\ref{figure:gallaghers_ic2} in Section~\ref{sec:ela}. It depicts the logarithm of the \textit{information sensitivity}~$\varepsilon$ against the \textit{(partial) information content}. Following these two curves -- the solid red line representing the \textit{information content}~$H(\varepsilon)$, and the dashed blue line representing the \textit{partial information content}~$M(\varepsilon)$ -- one can (more or less) easily derive the features from the plot. The blue diamond at the very left represents the \textit{initial partial information} $M_0$, the red circle on top of the solid, red line shows the \textit{maximum information content}~$H_{max}$, the green triangle indicates the \textit{ratio of partial information sensitivity}~$\varepsilon_{0.5}$ (with a ratio of $r = 0.5$) and the black square marks the \textit{settling sensitivity} $\varepsilon_s$ (with $s = 0.05$). Further details on the interpretation of this plot and the related features can be found in \cite{Munoz2012, Munoz2015}. In accordance to the code of the previous visualization techniques, such an information content plot can be created by the following command:

\begin{knitrout}
\definecolor{shadecolor}{rgb}{0.969, 0.969, 0.969}\color{fgcolor}\begin{kframe}
\begin{alltt}
\hlstd{R> }\hlkwd{plotInformationContent}\hlstd{(feat.object)}
\end{alltt}
\end{kframe}
\end{knitrout}

\begin{figure*}[!t]
  \centering
  \includegraphics[width = 0.72\textwidth, trim = 0mm 4mm 0mm 3mm, clip]{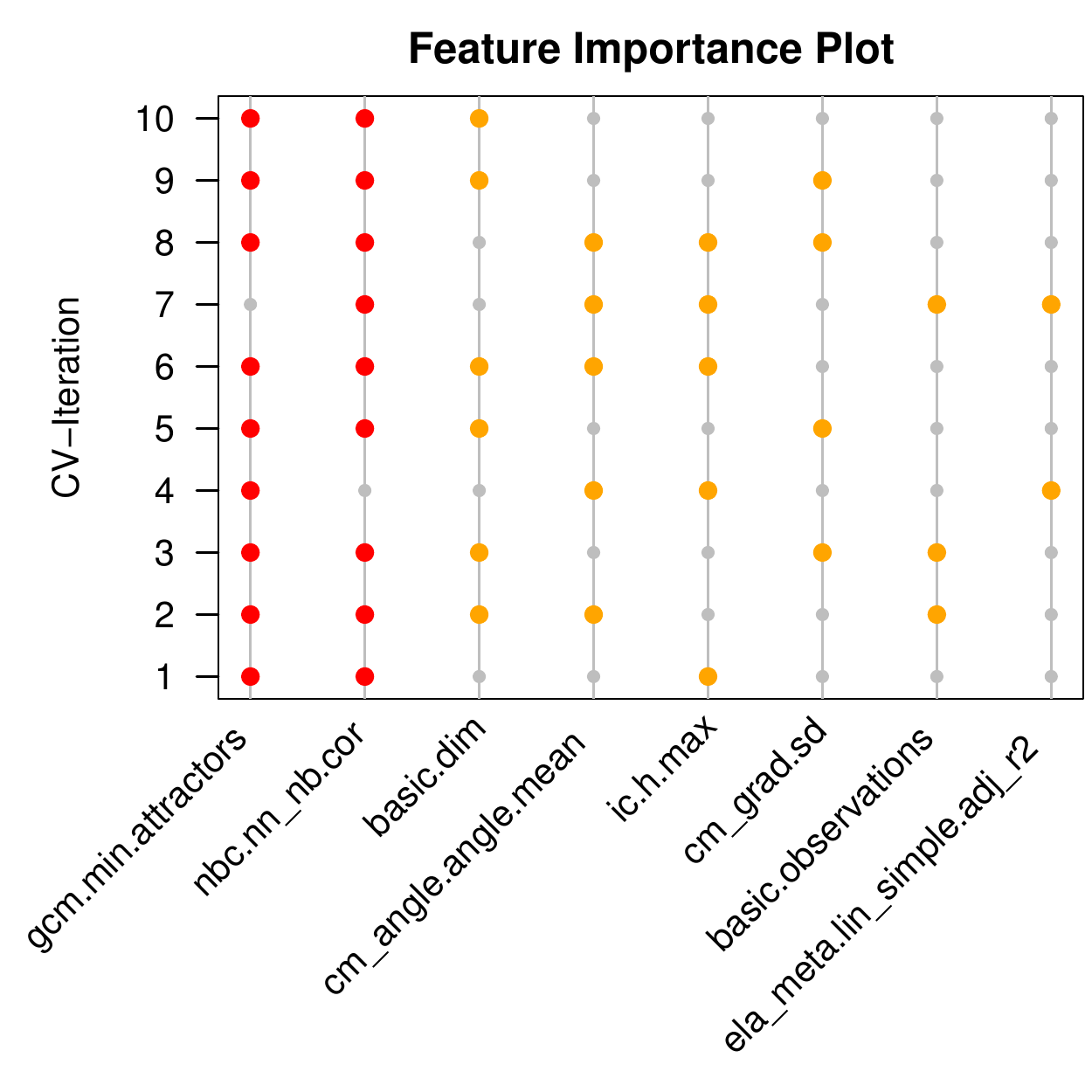}
	\caption{Plot of the feature importance as derived from a 10-fold cross-validated feature selection.}
	\label{figure:featimp}
\end{figure*}

In addition to the previously introduced visualization methods, \pkg{flacco} provides another plot function. However, in contrast to the previous ones, the \textit{Feature Importance Plot} (cf. Figure~\ref{figure:featimp}) does not visualize a specific feature set. Instead, it can be used to assess the importance of several features during a feature selection process of a machine learning (e.g.,~classification or regression) task. Given the high amount of features, provided by this package, it is very likely that many of them are redundant when training a (classification or regression) model and therefore, a feature selection strategy would be useful.

The scenario of Figure \ref{figure:featimp} shows the (artificially created) result of such a feature selection after performing a 10-fold cross validation using the R-package \pkg{mlr} \citep{Mlr2016}\footnote{A more detailed step-by-step example can be found in the documentation of the respective \pkg{flacco}-function \texttt{plotFeatureImportancePlot}.}. Such a resampling strategy is useful in order to assess whether a certain feature was selected because of its importance to the model or more or less just by chance. The red points show whether an important feature -- i.e.,~a feature, which has been selected in the majority of iterations (default: $\geq 80\%$) -- was selected within a specific iteration (= fold of the cross-validation). Less important features are depicted as orange points. The only information that is required to create such a \textit{Feature Importance Plot} is a list, whose elements (one per fold / iteration) are vectors of character strings. In the illustrated example, this list consists of ten elements (one per fold) with the first one being a character vector that contains exactly the three features \texttt{"gcm.min.attractors"}, \texttt{"nbc.nn\_nb.cor"} and \texttt{"ic.h.max"}. The entire list that leads to the plot from Figure \ref{figure:featimp} (denoted as \texttt{list.of.features}) would look like this:

\begin{knitrout}
\definecolor{shadecolor}{rgb}{0.969, 0.969, 0.969}\color{fgcolor}\begin{kframe}
\begin{alltt}
\hlstd{R> }\hlstd{list.of.features} \hlkwb{=} \hlkwd{readRDS}\hlstd{(}\hlstr{"list_of_features.rds"}\hlstd{)}
\hlstd{R> }\hlkwd{str}\hlstd{(list.of.features,} \hlkwc{width} \hlstd{=} \hlnum{77}\hlstd{,} \hlkwc{strict.width} \hlstd{=} \hlstr{"cut"}\hlstd{)}
\end{alltt}
\begin{verbatim}
List of 10
 $ : chr [1:3] "gcm.min.attractors" "ic.h.max" "nbc.nn_nb.cor"
 $ : chr [1:5] "gcm.min.attractors" "nbc.nn_nb.cor" "basic.dim" "cm_angle."..
 $ : chr [1:5] "gcm.min.attractors" "nbc.nn_nb.cor" "basic.dim" "basic.obs"..
 $ : chr [1:4] "gcm.min.attractors" "ic.h.max" "cm_angle.angle.mean" "ela_"..
 $ : chr [1:4] "gcm.min.attractors" "nbc.nn_nb.cor" "basic.dim" "cm_grad.sd"
 $ : chr [1:5] "gcm.min.attractors" "ic.h.max" "nbc.nn_nb.cor" "basic.dim" ..
 $ : chr [1:5] "ic.h.max" "nbc.nn_nb.cor" "cm_angle.angle.mean" "basic.obs"..
 $ : chr [1:5] "gcm.min.attractors" "ic.h.max" "nbc.nn_nb.cor" "cm_angle.a"..
 $ : chr [1:4] "gcm.min.attractors" "nbc.nn_nb.cor" "basic.dim" "cm_grad.sd"
 $ : chr [1:3] "gcm.min.attractors" "nbc.nn_nb.cor" "basic.dim"
\end{verbatim}
\end{kframe}
\end{knitrout}
\vspace*{-0.25cm}

Given such a list of features, the \textit{Feature Importance Plot} can be generated with the following command:

\vspace*{-0.2cm}
\begin{knitrout}
\definecolor{shadecolor}{rgb}{0.969, 0.969, 0.969}\color{fgcolor}\begin{kframe}
\begin{alltt}
\hlstd{R> }\hlkwd{plotFeatureImportance}\hlstd{(list.of.features)}
\end{alltt}
\end{kframe}
\end{knitrout}
\vspace*{-0.2cm}

Similarly to the computation of the features that has been described in Section~\ref{sec:examples}, each of the plots can be modified according to a user's needs by making use of the \texttt{control} parameter within each of the plot functions. Further details on the possible configurations of each of the visualization techniques can be found in their documentations.


\section{Graphical user interface}\label{sec:gui}

Within the previous sections, we have introduced \pkg{flacco}, which provides a wide collection of feature sets that were previously implemented in different programming languages or at least in different packages. Furthermore, we have shown how one can use it to compute the landscape features. While this is beneficial for researchers who are familiar with \proglang{R}, it comes with the drawback that researchers who are not familiar with \proglang{R} are left out. Therefore, \cite{Hanster2017} implemented a graphical user interface (GUI) for the package, which can either be started from within \proglang{R}, or -- which is probably more appealing to non-\proglang{R}-users -- as a platform-independent web-application.

The GUI was implemented using the \proglang{R}-package \pkg{shiny}~\citep{Shiny2016} and according to its developers, \pkg{shiny} \textit{``makes it incredibly easy to build interactive web applications with \proglang{R}''} and its \textit{``extensive pre-built widgets make it possible to build beautiful, responsive, and powerful applications with minimal effort.''}

If one wants to use the GUI-version that is integrated within \pkg{flacco} (version 1.5 or higher), one first needs to install \pkg{flacco} from CRAN (\url{https://cran.r-project.org/package=flacco}) and load it into the workspace of \proglang{R}.

\vspace*{-0.2cm}
\begin{knitrout}
\definecolor{shadecolor}{rgb}{0.969, 0.969, 0.969}\color{fgcolor}\begin{kframe}
\begin{alltt}
\hlstd{R> }\hlkwd{install.packages}\hlstd{(}\hlstr{"flacco"}\hlstd{,} \hlkwc{dependencies} \hlstd{=} \hlnum{TRUE}\hlstd{)}
\hlstd{R> }\hlkwd{library}\hlstd{(}\hlstr{"flacco"}\hlstd{)}
\end{alltt}
\end{kframe}
\end{knitrout}
\vspace*{-0.2cm}

Afterwards it only takes the following line of code to start the application:

\vspace*{-0.2cm}
\begin{knitrout}
\definecolor{shadecolor}{rgb}{0.969, 0.969, 0.969}\color{fgcolor}\begin{kframe}
\begin{alltt}
\hlstd{R> }\hlkwd{runFlaccoGUI}\hlstd{()}
\end{alltt}
\end{kframe}
\end{knitrout}
\vspace*{-0.35cm}

\newpage
If one rather prefers to use the platform-independent GUI, one can use the web-application, which is hosted on \url{https://flacco.shinyapps.io/flacco/}.

\begin{figure*}[!t]
  \centering
  \includegraphics[width = 0.715\textwidth, trim = 0mm 27.5mm 139.5mm 0mm, clip]{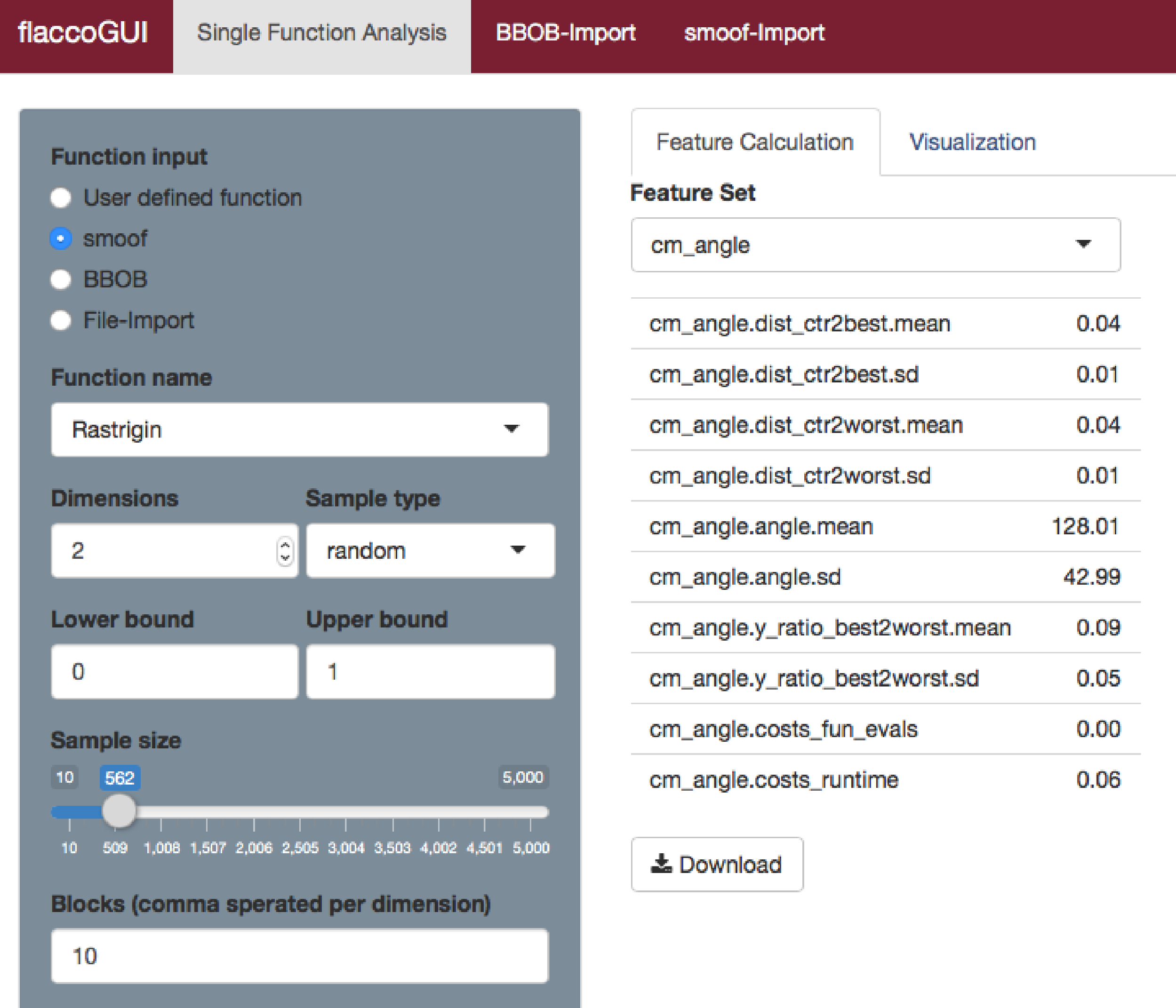}
  \caption{Screenshot of the GUI, after computing the \textit{cell mapping angle} features for a two-dimensional version of the \textit{Rastrigin} function as implemented in \pkg{smoof}.}
  \label{figure:gui_smoof}
\end{figure*}

\begin{figure*}[!p]
  \centering
  \includegraphics[width = \textwidth, trim = 0mm 5mm 0mm 1mm, clip]{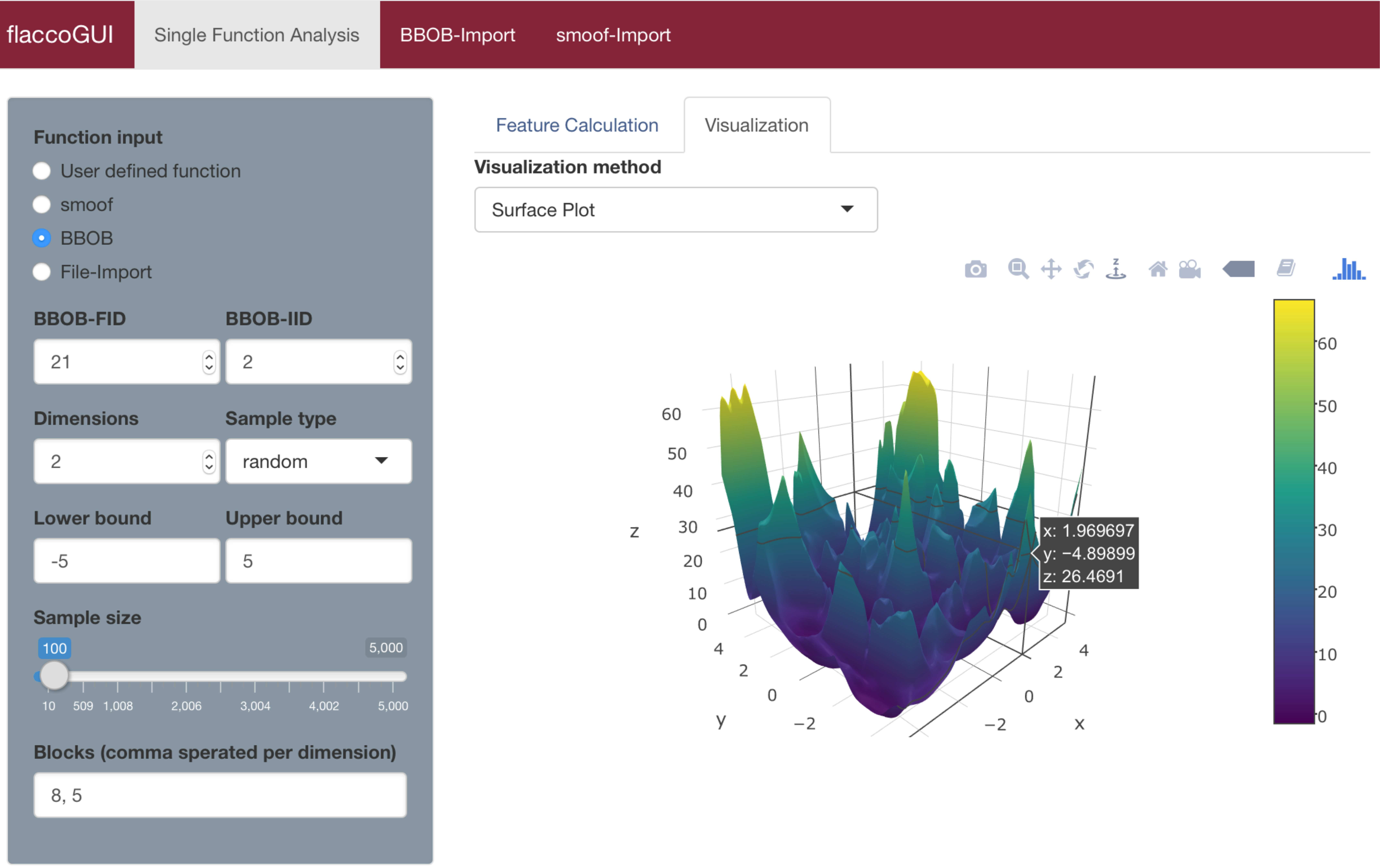}\\
  \vspace*{0.5cm}
  \includegraphics[width = \textwidth, trim = 0mm 12mm 0mm 1mm, clip]{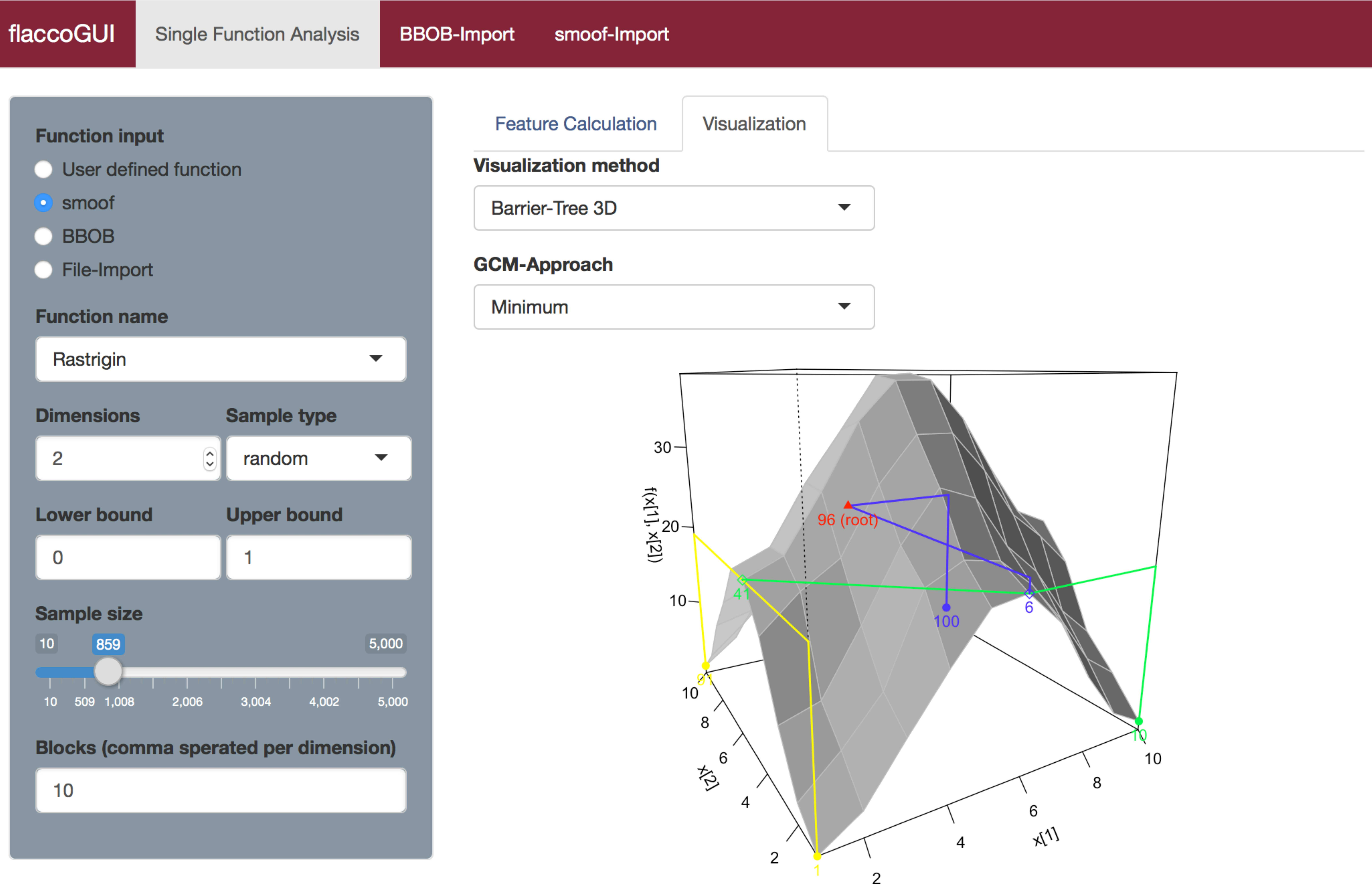}
  \caption{Exemplary screenshots of two visualization techniques as provided by the GUI and shown for two different optimization problems. Top: An interactive surface plot of an instance of \textit{Gallagher's 101-me Peaks} function, more precisely the second instance of the 21st BBOB problem, which for instance allows to zoom in and out of the landscape, as well as to rotate or shift it. Bottom: A 3-dimensional representation of the barrier tree for an instance of the \textit{Rastrigin} function as implemented in \pkg{smoof}.}
  \label{figure:gui_visual}
\end{figure*}

Once started, the application shows a bar on the top, where the user can select between the following three options (as depicted in Figure~\ref{figure:gui_smoof}): ``Single Function Analysis'', ``BBOB-Import'' and ``Smoof-Import''. Selecting the first tab, the application will show two windows: (1) a box (colored in grey) on the left, which inquires all the information that is needed for creating the feature object, i.e., the underlying function, the problem dimension (= number of input variables), the boundaries, the size and sampling type (latin hypercube or random uniform sampling) of the initial sample and the number of cells per dimension, and (2) a screen for the output on the right, where one can either compute the numerical landscape features under the tab ``Feature Calculation'' (as shown in Figure~\ref{figure:gui_smoof}) or create a related plot under the tab ``Visualization'' (as shown in Figure~\ref{figure:gui_visual}).

For the definition of the underlying function (within the grey box), the user has the following options: (1) providing a user-defined function by entering the corresponding \proglang{R} expression, e.g.,~\texttt{sum(x\^}\texttt{2)}, into a text box, (2) selecting one of the single-objective problems provided by \pkg{smoof} from a drop-down menu, (3) defining a BBOB function~\citep{Hansen2010} via its function ID (FID) and instance ID (IID), or (4) ignoring the function input and just provide the initial design, i.e., the input variables and the corresponding objective value, by importing a CSV-file.

Immediately after the necessary information for the feature object is provided, the application will automatically produce the selected output on the right half of the screen. If one does not select anything (on the right half), the application will try to compute the \textit{cell mapping angle} features, as this is the default selection at the start of the GUI. However, the user can select any other feature set, as well as the option ``all features'', from a drop-down menu. Once the computation of the features is completed, one can download the results as a CSV-file by clicking on the ``Download''-button underneath the feature values.

Aside from the computation of the landscape features, the GUI also provides functionalities for generating multiple visualizations of the given feature object. Aside from the \textit{cell mapping plots} and two- and three-dimensional \textit{barrier trees}, which can only be created for two-dimensional problems, it is also able to produce \textit{information content plots} (cf.~Section~\ref{sec:visual}). Note that the latter currently is the only plotting function that is able to visualize problems with more than two input variables.
In addition to the aforementioned plots, the application also enables the visualization of the function itself -- if applicable -- by means of an interactive \textit{function plot} (for one-dimensional problems), as well as two-dimensional \textit{contour} or three-dimensional \textit{surface plots} (for two-dimensional problems). The interactivity of these plots, such as the one that is shown in the upper image of Figure~\ref{figure:gui_visual}, was provided by the \proglang{R}-package \pkg{plotly}~\citep{Plotly2016} and (amongst others) enables to zoom in and out of the landscape, as well as to rotate or shift the latter.

While \pkg{flacco} itself allows the configuration of all of its parameters -- i.e., the parameters used by any of the feature sets, as well as the ones related to the plotting functions -- the GUI provides a simplified version of this. Hence, for the cell mapping and barrier tree plots, one can choose the approach for selecting the representative objective value per cell: (a) the best objective value among the samples from the cell (as shown in the lower image of Figure~\ref{figure:gui_visual}), (b) the average of all objective values from the cell's sample points, or (c) the objective value of the observation that is located closest to the cell's center. In case of the information content plot, one can configure the range of the x-axis, i.e., the range of the (logarithmic) values of the information sensitivity parameter $\varepsilon$.

The previous paragraphs described the computation and visualization when dealing with a single optimization problem. However, as already mentioned in Section~\ref{sec:examples}, the authors highly recommend not to interpret single features on single optimization problems. Instead, one should rather compute several (sets of) landscape features across an entire benchmark of optimization problems. This way, they might be able to provide the information for distinguishing the benchmark's problems from each other and predicting a well-performing optimization algorithm per instance -- cf.~the \textit{Algorithm Selection Problem}, which was already mentioned as a potential use case in Section~\ref{sec:intro} -- or for adapting the parameters of such an optimizer to the underlying landscape.

\begin{figure*}[!t]
  \centering
  \includegraphics[width = \textwidth, trim = 0mm 13mm 0mm 4mm, clip]{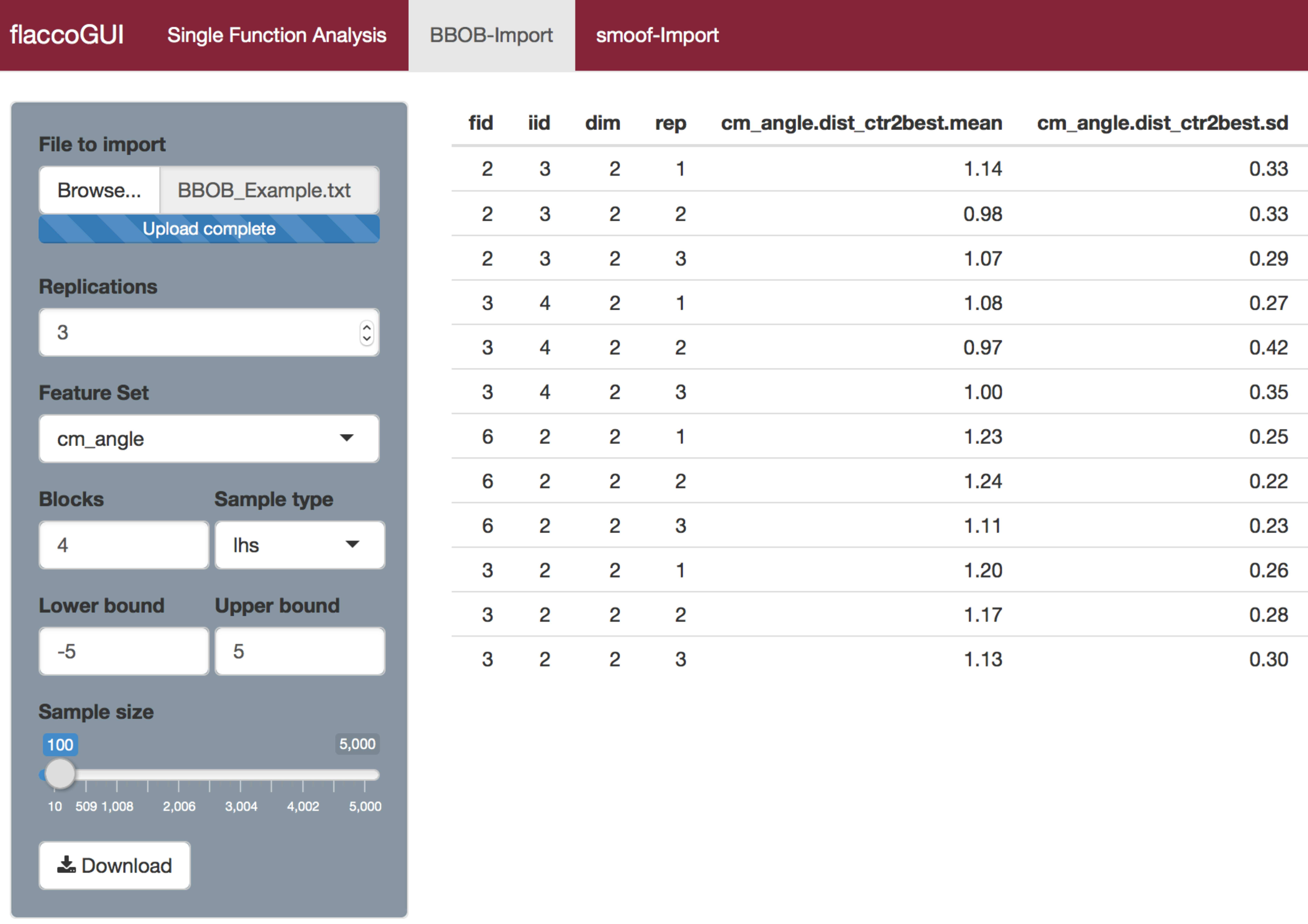}
  \caption{Screenshot of the GUI, after computing the \textit{cell mapping angle} features for four different BBOB instances with three replications each. Note that the remaining eight features from this set were computed as well and were simply cut off for better visibility.}
  \label{figure:gui_bbob}
\end{figure*}

In order to facilitate such experiments, our application also provides functionalities for computing a specific feature set (or even all feature sets) simultaneously on multiple instances of a specific problem class -- such as the BBOB functions or any of the other single-objective optimization problems provided by \pkg{smoof}. Figure~\ref{figure:gui_bbob} shows an exemplary screenshot of how such a batch computation could look like for the BBOB problems. In a first step, one has to upload a CSV-file consisting of three columns -- the function ID (FID), instance ID (IID) and problem dimension (dim) -- and one row per problem instance. Next, one defines the desired number of replications, i.e., how often should the features be computed per instance\footnote{As many of the features are stochastic, it is highly recommended to compute the features multiple (= at least 5 to 10) times.}. Afterwards, one selects the feature set that is supposed to be computed, defines number of blocks and boundaries per dimension, as well as the desired sample type and size. Note that, depending on the size of the benchmark, i.e., the number of instances and replications, the number and dimensions of the initial sample, as well as the complexity of the chosen feature set, the computation of the features might take a while. Furthermore, please note that the entire application is reactive. That is, once all the required fields provide some information, the computation of the features will start and each time the user changes the information within a field, the computation will automatically restart. Therefore, the authors recommend to first configure all the parameters prior to uploading the CSV-file as the latter is the only missing piece of information at the start of the application. Once the feature computation was successfully completed, one can download the computed features / feature sets as a CSV-file by clicking on the ``Download''-Button.

Analogously to the batch-wise feature computation on BBOB instances, one can also perform batch-wise feature computation for any of the other single-objective problems that are implemented in \pkg{smoof}. For this purpose, one simply has to go to the tab labelled ``smoof-Import'' and perform the same steps as described before. The only difference is, that the CSV-file with the parameters now only consists of a single column (with the problem dimension).


\section{Summary}\label{sec:summary}

The \proglang{R}-package \pkg{flacco} provides a collection of numerous features for exploring and characterizing the landscape of continuous, single-objective (optimization) problems by means of numerical values, which can be computed based on rather small samples from the decision space. Having this wide variety of feature sets bundled within a single package simplifies further researches significantly --~especially in the fields of algorithm selection and configuration -- as one does not have to run experiments across multiple platforms. In addition, the package comes with different visualization strategies, which can be used for analyzing and illustrating certain feature sets, leading towards an improved understanding of the features.

Furthermore, this package can be meaningfully combined with other \proglang{R}-packages such as \pkg{mlr} \citep{Mlr2016} or \pkg{smoof} \citep{Smoof2016} in order to achieve better results when solving algorithm selection problems. Consequently, \pkg{flacco} should be seen as a useful toolbox when performing Exploratory Landscape Analysis and when working towards algorithm selection models of single-objective, continuous optimization problems. Therefore, the package can be considered as a solid foundation for further researches -- as for instance performed by members of the COSEAL group\footnote{COSEAL is an international group of researchers with a focus on the \textbf{Co}nfiguration and \textbf{Se}lection of \textbf{Al}gorithms, cf. \url{http://www.coseal.net/}} -- allowing to get a better understanding of algorithm selection scenarios and the employed optimization algorithms.

Lastly, \pkg{flacco} also comes with a graphical user interface, which can either be run from within \proglang{R}, or alternatively as a platform-independent standalone web-application. Therefore, even people who are not familiar with \proglang{R} have the chance to benefit of the functionalities of this package -- especially from the computation of numerous landscape features -- without the burden of interfacing or re-implementing identical features in a different programming language.


\section*{Acknowledgements}
The author acknowledges support by the European Center of Information Systems~(ERCIS)\footnote{The ERCIS is an international network in the field of Information Systems, cf. \url{https://www.ercis.org/}}. He also appreciates the support from Carlos Hern{\'a}ndez (from CINVESTAV in Mexico City, Mexico) and Jan Dagef{\"o}rde (from the University of M{\"u}nster, Germany) when translating the \proglang{MATLAB}-code of some feature sets into \proglang{R}-code. Furthermore, the author thanks Christian Hanster for his valuable contributions to the package's GUI.


\bibliography{jss_flacco_arxiv}

\end{document}